\documentclass[letterpaper, 10 pt, conference]{ieeeconf}  

\IEEEoverridecommandlockouts                              

\usepackage{graphics} 
\usepackage{epsfig} 
\usepackage{mathptmx} 
\usepackage{times} 
\usepackage{amsmath} 
\usepackage{amssymb}  
\usepackage{pgfplots}
 \usepackage{multirow}
\usepackage{bm}
\usepackage{subcaption}
\usepackage{authblk}
\usepackage{adjustbox}

\title{\LARGE \bf
MGSO: Monocular Real-time Photometric SLAM with Efficient 3D Gaussian Splatting
}

\author{Yan Song Hu$^{1}$, Nicolas Abboud$^{1,2}$, Muhammad Qasim Ali$^{1}$, Adam Srebrnjak Yang$^{1}$,
Imad Elhajj$^{2}$, \\Daniel Asmar$^{2}$, Yuhao Chen$^{1}$, John S. Zelek$^{1}$
\thanks{$^{1}$Yan Song Hu, N. Abboud, M. Ali, A. Yang, Y. Chen, and J. Zelek are with Vision and Image Processing Lab at the Faculty of System Design Engineering, at the University of Waterloo, 200 University Avenue West, Waterloo, Ontario, Canada; emails: {\tt\small y324hu,n2abboud,m45ali,asyang,yuhao.chen1, jzelek@uwaterloo.ca}
}%
\thanks{$^{2}$ N. Abboud, I. Elhajj and Daniel Asmar are with Vision and Robotics Lab, Maroun Semaan Faculty of Engineering and Architecture, American University of Beirut, 1107 2020, Riad El Solh, Beirut,
Lebanon; emails: {\tt\small nfa53,ie05,da20@aub.edu.lb} }
}

\begin{document}
\makeatletter
\let\@oldmaketitle\@maketitle
\renewcommand{\@maketitle}{\@oldmaketitle
\vspace{0.1cm} 
\centering
\includegraphics[height=4.375cm]{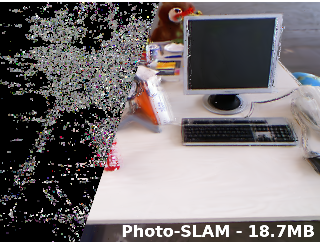}
\includegraphics[height=4.375cm]{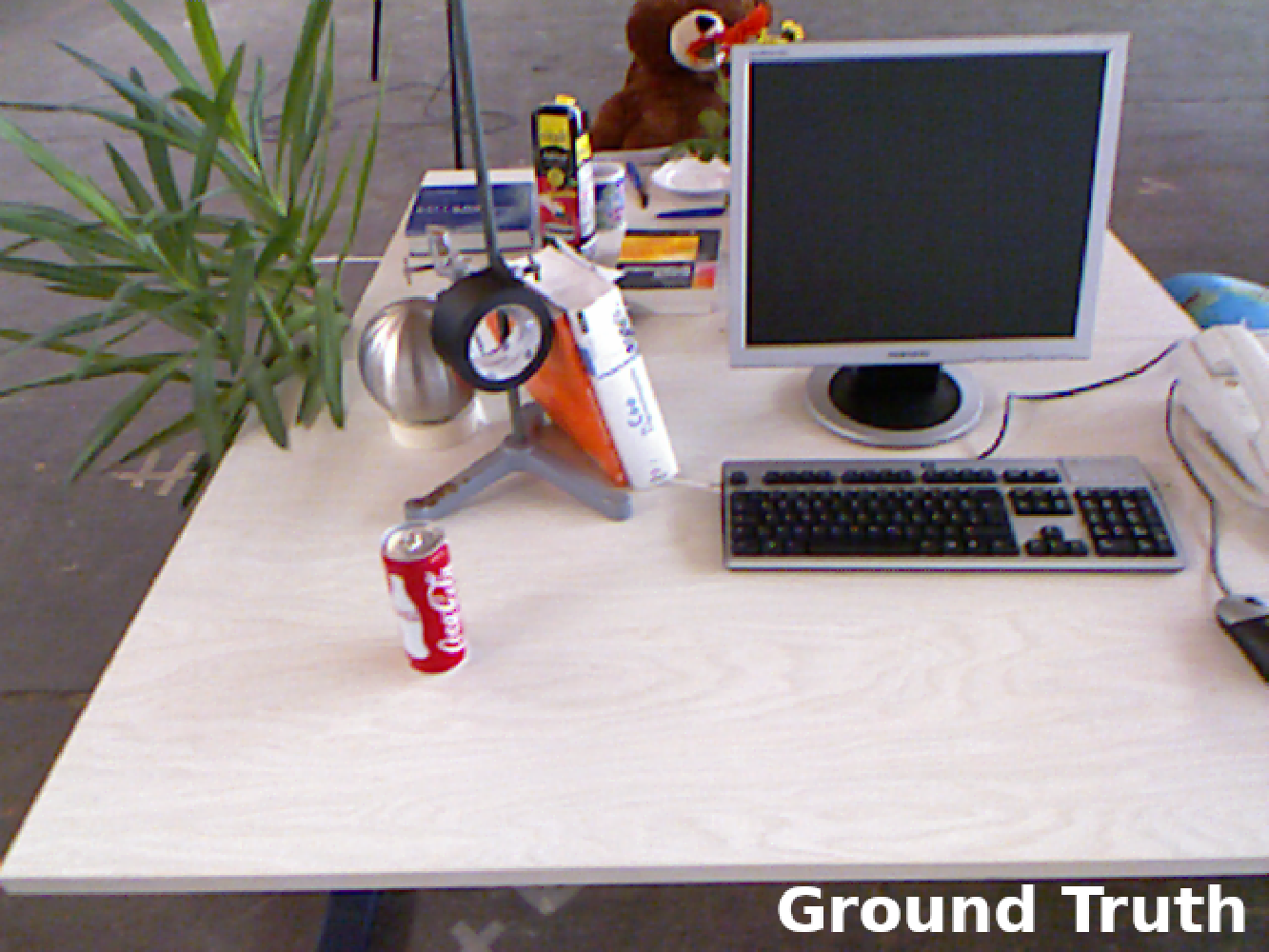}
\includegraphics[height=4.375cm]{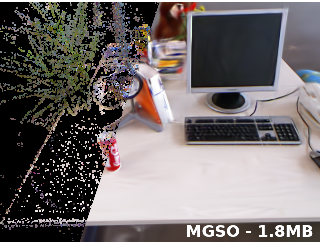}
\captionof{figure}{Qualitative renders of the TUM-RGBD dataset \cite{sturm12iros} with input point clouds. By initializing 3D Gaussian Splatting (3DGS) \cite{kerbl3Dgaussians} with dense, structured point clouds, MGSO produces reconstructions that are memory-efficient and high quality.}

\label{fig:gsoheader}}
\makeatother

\maketitle
\thispagestyle{empty}
\pagestyle{empty}


\begin{abstract}
Real-time SLAM with dense 3D mapping is computationally challenging, especially on resource-limited devices. The recent development of 3D Gaussian Splatting (3DGS) offers a promising approach for real-time dense 3D reconstruction. However, existing 3DGS-based SLAM systems struggle to balance hardware simplicity, speed, and map quality. Most systems excel in one or two of the aforementioned aspects but rarely achieve all. A key issue is the difficulty of initializing 3D Gaussians while concurrently conducting SLAM. To address these challenges, we present Monocular GSO (MGSO), a novel real-time SLAM system that integrates photometric SLAM with 3DGS. Photometric SLAM provides dense structured point clouds for 3DGS initialization, accelerating optimization and producing more efficient maps with fewer Gaussians. As a result, experiments show that our system generates reconstructions with a balance of quality, memory efficiency, and speed that outperforms the state-of-the-art. Furthermore, our system achieves all results using RGB inputs. We evaluate the Replica, TUM-RGBD, and EuRoC datasets against current live dense reconstruction systems. Not only do we surpass contemporary systems, but experiments also show that we maintain our performance on laptop hardware, making it a practical solution for robotics, A/R, and other real-time applications.

\end{abstract}

\section{Introduction}








Simultaneous localization and mapping (SLAM) is a fundamental task in autonomous robot navigation. It is the process by which a robot constructs a map of an environment while concurrently keeping track of its own location. Accurate self-localization is an essential precursor for advanced mobile robot tasks. Traditionally, SLAM systems provide semantically-poor map representations that are efficient for localization and basic navigation but lack the details needed for complex tasks. For example, sparse point clouds are efficient for localization but lack the surface detail needed for robotic grasping. For these complex robotic tasks, dense, high-fidelity spatial data is increasingly important.

To meet this demand, SLAM systems have evolved to generate dense 3D maps while still simultaneously performing localization. Dense SLAM systems are categorized into two approaches: decoupled and coupled. Decoupled approaches separate tracking from reconstruction, using a traditional SLAM system to provide outputs for a dense reconstruction process. Coupled approaches integrate dense reconstruction with both mapping and tracking, improving map quality but often facing speed bottlenecks, as accurate localization depends on building a high-quality map, which takes time.

A key challenge in decoupled systems is the lack of synergy between SLAM and dense reconstruction components. SLAM algorithms often fail to provide optimal data for high-quality dense reconstruction, compromising overall system performance. To address this challenge, we tailored our SLAM system to meet the specific needs of 3D Gaussian Splatting (3DGS) \cite{kerbl3Dgaussians}. 3DGS typically requires an initial point cloud to begin reconstruction, with denser, well-structured initial point clouds leading to improved and faster results \cite{hu2024realtimegaussiansplattingaccelerating}. However, traditional feature-based SLAM methods produce sparse point clouds that are not optimal for 3DGS initialization. While RGB-D data could provide dense and accurate point clouds, using a monocular camera is preferable for wider applicability.

In this paper, we introduce \textbf{Monocular-GSO (MGSO)}, a dense visual SLAM system that performs high-quality online 3D reconstruction in real-time using a single monocular camera. MGSO is a decoupled system that employs photometric SLAM to initialize a 3D Gaussian Splatting (3DGS) module running in parallel, enabling live dense scene reconstruction. The MGSO acronym is a blend of Direct Sparse Odometry (DSO) \cite{dso:engel18}, the photometric SLAM system we built upon, and Gaussian Splatting (GS). In contrast to conventional feature-based SLAM methods that generate sparse point clouds, MGSO is designed to track a dense set of pixels, yielding a denser and well-structured point cloud output. We leverage this dense, structured point cloud to initialize 3D Gaussian Splatting (3DGS) in unmapped areas. Initializing with a high-quality set of points accelerates 3DGS optimization, guiding it toward more compact reconstructions with fewer artifacts and redundancies. As a result, our approach leads to real-time reconstruction with dense 3D maps with high quality and memory compactness. 

The main contributions of MGSO are as follows:
\begin{itemize}
    \item A real-time dense SLAM system that harnesses the synergy between photometric SLAM and 3DGS.
    \item Our system only requires a monocular camera.
    \item Experiments show that our system has a combination of speed, map quality, and memory efficiency unmatched by other dense SLAM systems.
\end{itemize}

\section{Related Work}



Dense SLAM research has long explored various 3D representations such as signed distance functions \cite{izadi2011kinectfusion}, dense point clouds \cite{densergd:kerl2013}, and surfel clouds \cite{badslam:schops2019}\cite{whelan2016elasticfusion}. Despite these advancements, efficiently generating high-quality maps for real-time applications remains challenging. Recent innovations in Neural Radiance Fields (NeRFs) \cite{mildenhall2020nerf} and 3D Gaussian Splatting \cite{kerbl3Dgaussians} show promise in addressing the issue. These approaches offer high-quality representations that are easy to create and render. Consequently, this section will focus on systems based on these two techniques.

NeRFs represent scenes using a neural network that outputs novel views based on the input camera's position and rotation. They also allow for the incremental learning and updating of their 3D representations through gradient-based optimization \cite{mildenhall2020nerf}. This capability has been effectively applied in pioneering works like iMap \cite{imap:Sucaretal:ICCV2021}, and further improved by subsequent systems such as NICE-SLAM \cite{Zhu2022CVPR}, Orbeez-SLAM \cite{10160950}, and NeRF-SLAM \cite{10341922}. However, NeRF-based SLAM systems face two notable challenges: they require predefined scene bounds, which is often impractical in exploratory environments; and their implicit scene representations can be difficult to integrate with other systems. These limitations have spurred the adoption of 3DGS as a more suitable dense reconstructor.

3DGS is an approach to scene representation that models the environment as a large set of 3D Gaussians, which resemble blurry overlapping clouds. Rendering images involves projecting Gaussians onto the camera plane, depth-sorting, and blending them front-to-back. Similar to NeRF, 3DGS allows for gradient-based optimization of parameters by minimizing the discrepancy between rendered and input images. The method will also heuristically clone or prune Gaussians over time. 3DGS offers a boundless and fast-to-render representation compared to NeRF, making it highly suitable for real-time SLAM applications. 

Early 3DGS-based SLAM systems, like MonoGS \cite{Matsuki:Murai:etal:CVPR2024}, SplaTAM \cite{keetha2024splatam}, GS-SLAM \cite{yan2023gs}, and Gaussian-SLAM \cite{yugay2023gaussianslam}, utilize a one-stage approach where tracking and mapping are tightly coupled. This approach introduces a dependency on map refinement before tracking can proceed, which results in slow performance, as shown in Table \ref{table1}. Even newer coupled systems such as CG-SLAM \cite{hu2024cg}, RTG-SLAM \cite{peng2024rtg}, and SplatSLAM \cite{sandström2024splatslamgloballyoptimizedrgbonly} struggle to run at speeds faster than 20 fps. To allow dense 3DGS-SLAM to operate faster, two-stage systems like Photo-SLAM \cite{hhuang2024photoslam}, IG-SLAM \cite{sarikamis2024igslam} and GS-ICP \cite{Ha2024RgbdGS} emerged, decoupling the tracking and mapping functions. Furthermore, the majority of current 3DGS systems heavily rely on depth data to perform 3D reconstruction (Table \ref{table1}), making them dependant on RGB-D sensors.

\begin{table}[htbp]
\caption{Existing 3DGS SLAM Systems}
\begin{center}
\begin{tabular}{c|c c c}
\hline
Name & Type & Possible Sensors & FPS  \\
\hline
MonoGS \cite{Matsuki:Murai:etal:CVPR2024} & Coupled & RGB,RGB-D & \textless 5\\
SplaTAM \cite{keetha2024splatam} & Coupled & RGB-D & \textless 5\\
GS-SLAM \cite{yan2023gs} & Coupled & RGB-D & \textgreater 5, \textless 10\\
Gaussian-SLAM \cite{yugay2023gaussianslam} & Coupled & RGB-D & \textless 5\\
CG-SLAM \cite{hu2024cg}& Coupled & RGB-D & \textgreater 15, \textless 20\\
RTG-SLAM \cite{peng2024rtg}& Coupled & RGB-D & \textgreater 15, \textless 20\\
SplatSLAM \cite{sandström2024splatslamgloballyoptimizedrgbonly}& Coupled & RGB & \textless 5\\
\hline
GS-ICP \cite{Ha2024RgbdGS}& Decoupled & RGB-D & \textgreater 30\\
Photo-SLAM \cite{hhuang2024photoslam}& Decoupled & RGB*,RGB-D & \textgreater 30\\
IG-SLAM \cite{sarikamis2024igslam}& Decoupled & RGB & \textgreater 5, \textless 10\\
MGSO (ours) & Decoupled & RGB & \textgreater 30\\
\hline
\multicolumn{4}{l}{*Both monocular and stereo}
\end{tabular}
\label{table1}
\end{center}
\end{table}

Our system, MGSO, improves on existing two-stage 3DGS-based SLAM systems while utilizing only RGB data. It operates at 30 fps or higher, a performance matched only by Photo-SLAM and GS-ICP (see Table \ref{table1}). While MGSO is most similar to Photo-SLAM, which combines 3DGS with ORBSLAM3, we address Photo-SLAM's tendency to create large, memory-inefficient maps. GS-ICP offers exceptional speed but requires depth data to initialize its iterative closest point tracking, whereas our system operates using only RGB data. Unlike IG-SLAM \cite{sarikamis2024igslam}, which uses pseudo-depth RGB-D data at the cost of slower performance, MGSO maintains real-time speeds while generating accurate, compact maps. 



\section{Method}



MGSO integrates two core components that operate concurrently: a SLAM module responsible for accurate pose estimation, and a 3D dense reconstruction module for mapping. 


\subsection{SLAM module:}
The tracking backbone of our system is built upon a lineage of visual SLAM approaches originating from Direct Sparse Odometry (DSO) \cite{dso:engel18}. DSO's key innovation is  demonstrating that selective pixel sampling for photometric tracking enhances localization accuracy compared to using all pixels in an image. We chose to build upon DSO because we found that its pixel selection strategy also aligns well with initializing 3DGS. 
DSO tracks a set of pixels across consecutive frames \(i\) and \(j\), optimizing the camera pose (\(\bm{p}\)) by minimizing the photometric loss equation below for each pixel tracked,

\newcommand{\norm}[1]{\left\lVert#1\right\rVert}
\begin{equation}
E=\norm{
    (\mathit{I}_j[\bm{p}_j]-b_j) -
    \frac{s_j a_j}{s_i a_i}
    (\mathit{I}_i[\bm{p}_i]-b_i)
}
\end{equation}


\noindent where \(I\) queries the pixel intensity, \(a\) and \(b\) are variables to account for lighting changes, and \(s\) is the camera exposure. The basic principle of this loss equation is to identify pose changes that best match the pixel intensity variation between consecutive frames \(i\) and \(j\). The equation is applied at both tracking and mapping levels.

At each frame, our system's tracking process calculates pose changes relative to the latest keyframe, assuming a fixed map. The map of tracked pixels is only adjusted when a keyframe is inserted. A new keyframe is a reference frame that captures a distinct view of the scene relative to existing keyframes. When mapping is done, all current keyframe poses and the map, which consists of the tracked pixel points, are adjusted. Our system then converts the map of tracked pixels into a point cloud map and adds it, along with the keyframe poses, to the dense reconstruction module. We adopt DSO's windowed keyframe management strategy, which generates keyframes when significant changes in the field of view, rotation, or lighting are detected. Older keyframes are removed if the number of keyframes exceeds the window size, which by default is eight, using a distance-based score to ensure a well-distributed set of keyframes in 3D space.

\begin{figure}[hpb]
    \centering
    {\renewcommand{\arraystretch}{0}
    \begin{tabular}{c@{}c}
    \begin{subfigure}[b]{.485\columnwidth}
        \centering
    \includegraphics[width=\columnwidth]{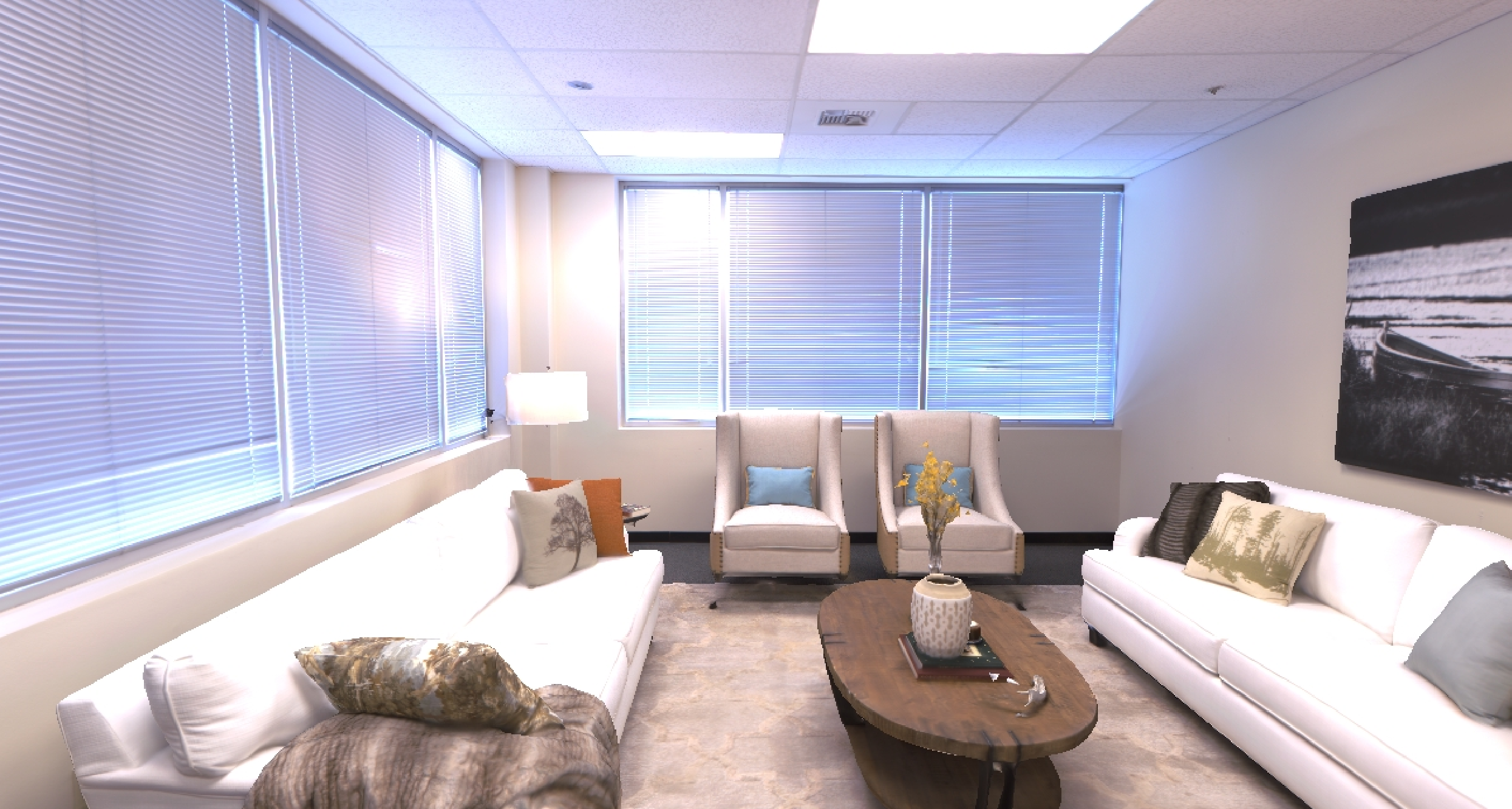}%
    \end{subfigure}&
    \begin{subfigure}[b]{.485\columnwidth}  
        \centering
    \includegraphics[width=\columnwidth]{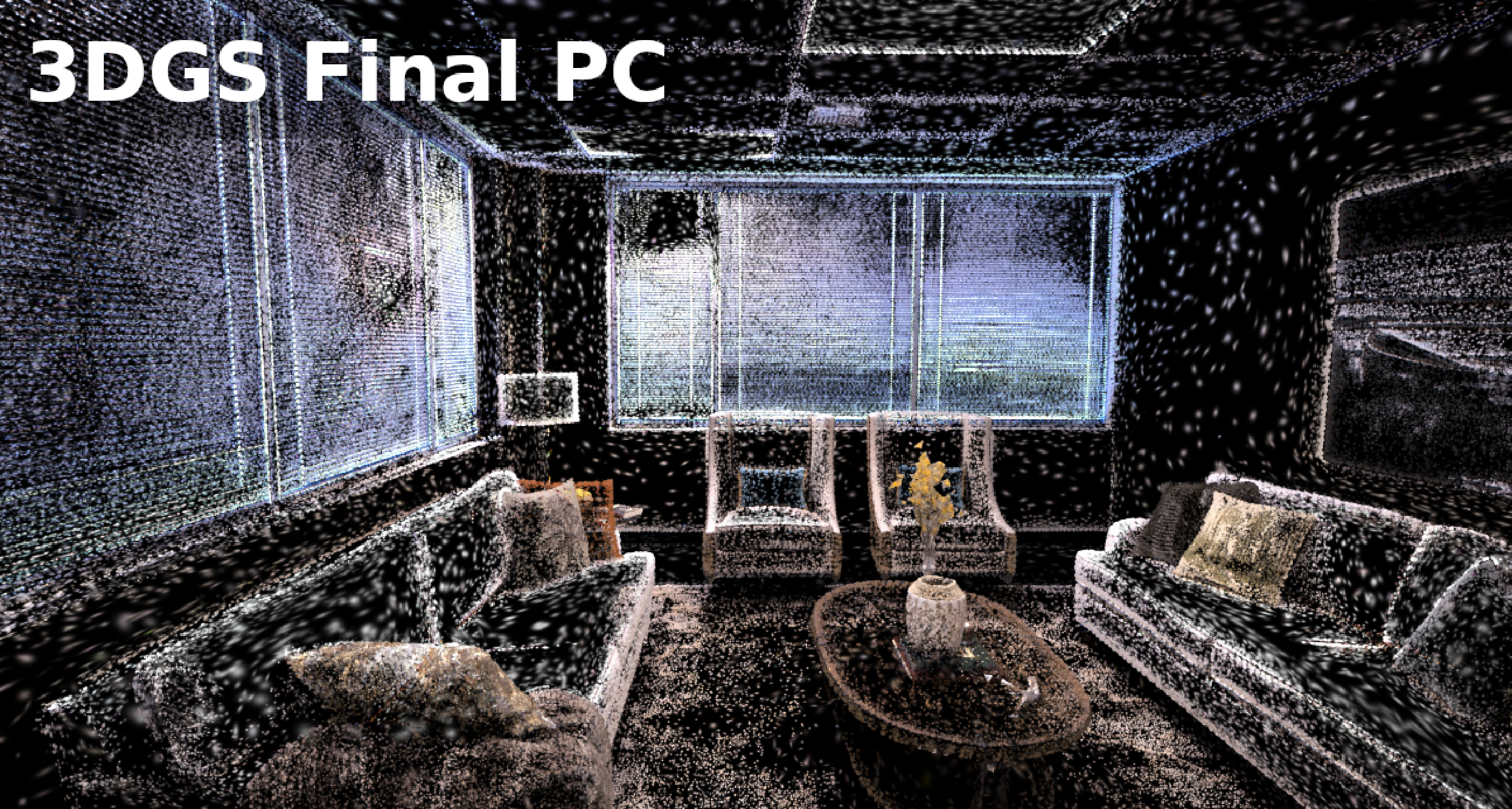}
    \end{subfigure}\\
    \begin{subfigure}[t]{.485\columnwidth}   
        \centering 
    \includegraphics[width=\textwidth]{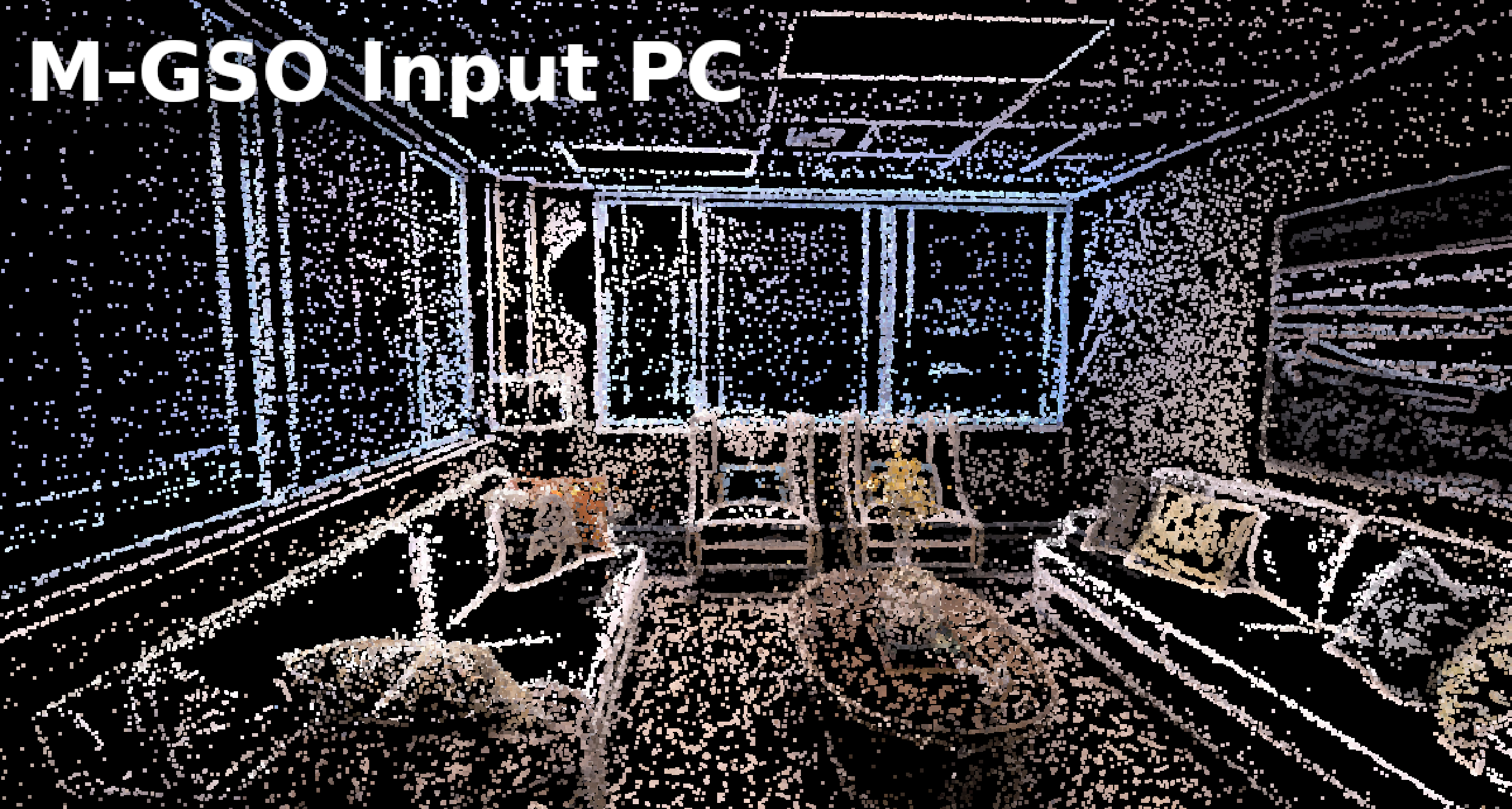}%
    \end{subfigure}&
    \begin{subfigure}[t]{.485\columnwidth}   
        \centering 
        \includegraphics[width=\columnwidth]{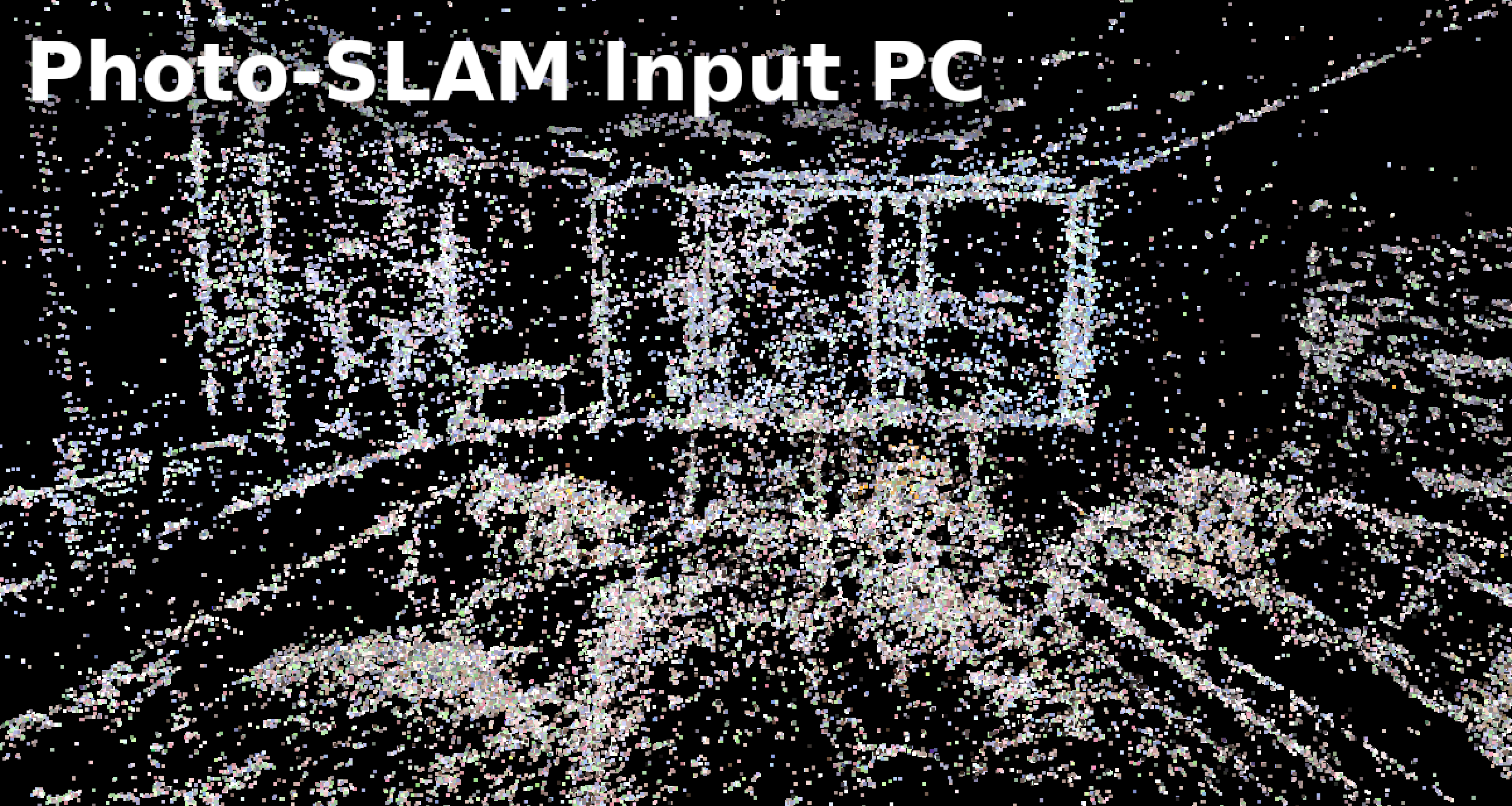}%
    \end{subfigure}
    \end{tabular}}
      \caption{Comparison of 3DGS point clouds from MGSO and Photo-SLAM on Replica room0. Top left: original frame; top right: Map from original 3DGS after 10,240 iterations with Gaussian size set to 0.1. Bottom: MGSO vs. Photo-SLAM point clouds.}
      \label{comparefigures}
   \end{figure}

The inspiration for our method is from analyzing the final 3DGS Gaussian position of the original 3DGS (Figure \ref{comparefigures}). We realized that the final position, colour, and distribution of Gaussians of the final map resembled the point cloud output from DSO (Figure \ref{comparefigures}). From this observation, we conjectured that initializing 3DGS with photometric SLAM would enhance 3DGS optimization because it would reduce the required optimizations. 

\begin{figure}[hpb]
\centering
\begin{subfigure}{.245\textwidth}
  \centering
  \includegraphics[width=0.95\linewidth]{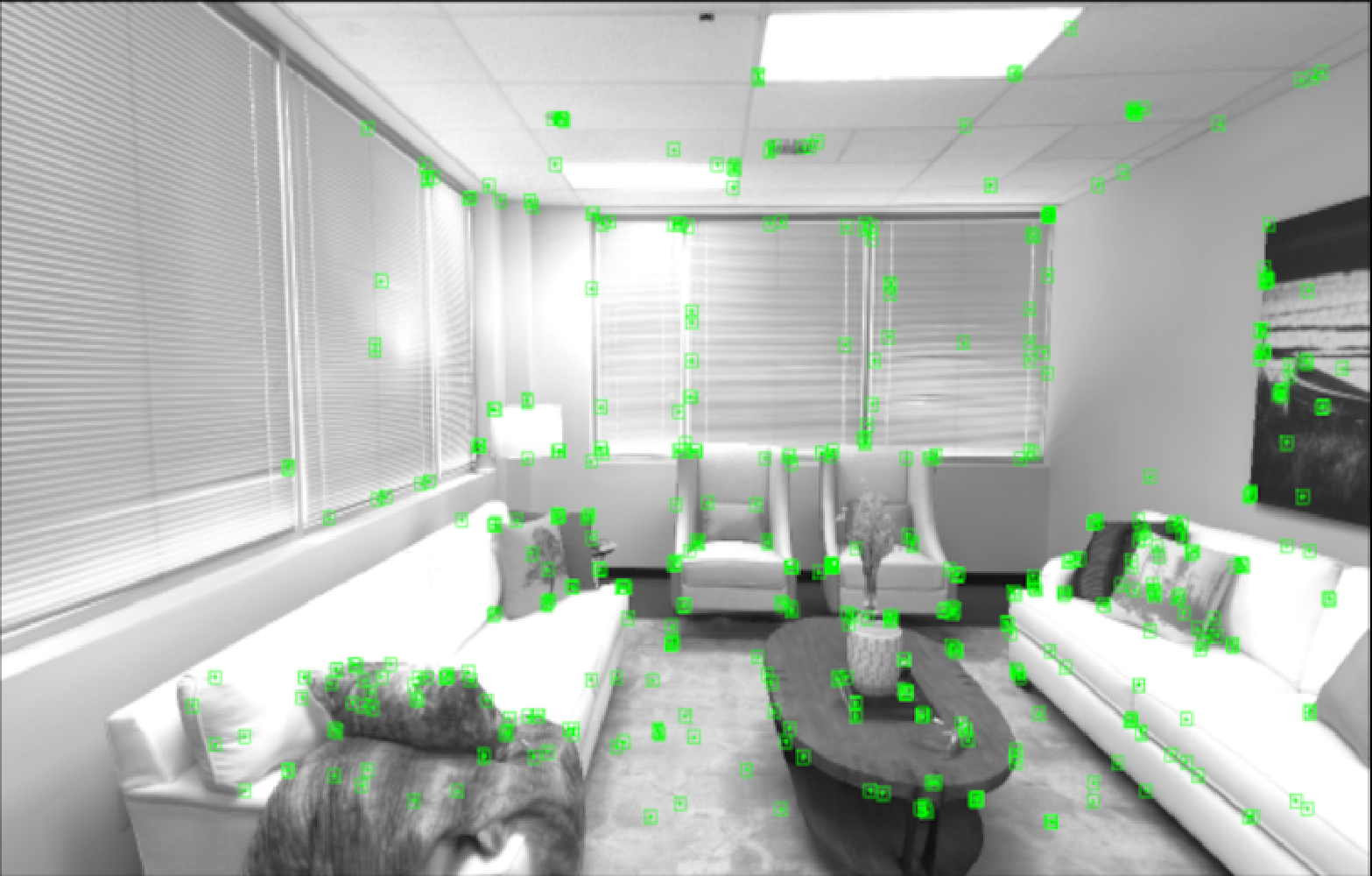}
\end{subfigure}%
\begin{subfigure}{.245\textwidth}
  \centering
  \includegraphics[width=0.95\linewidth]{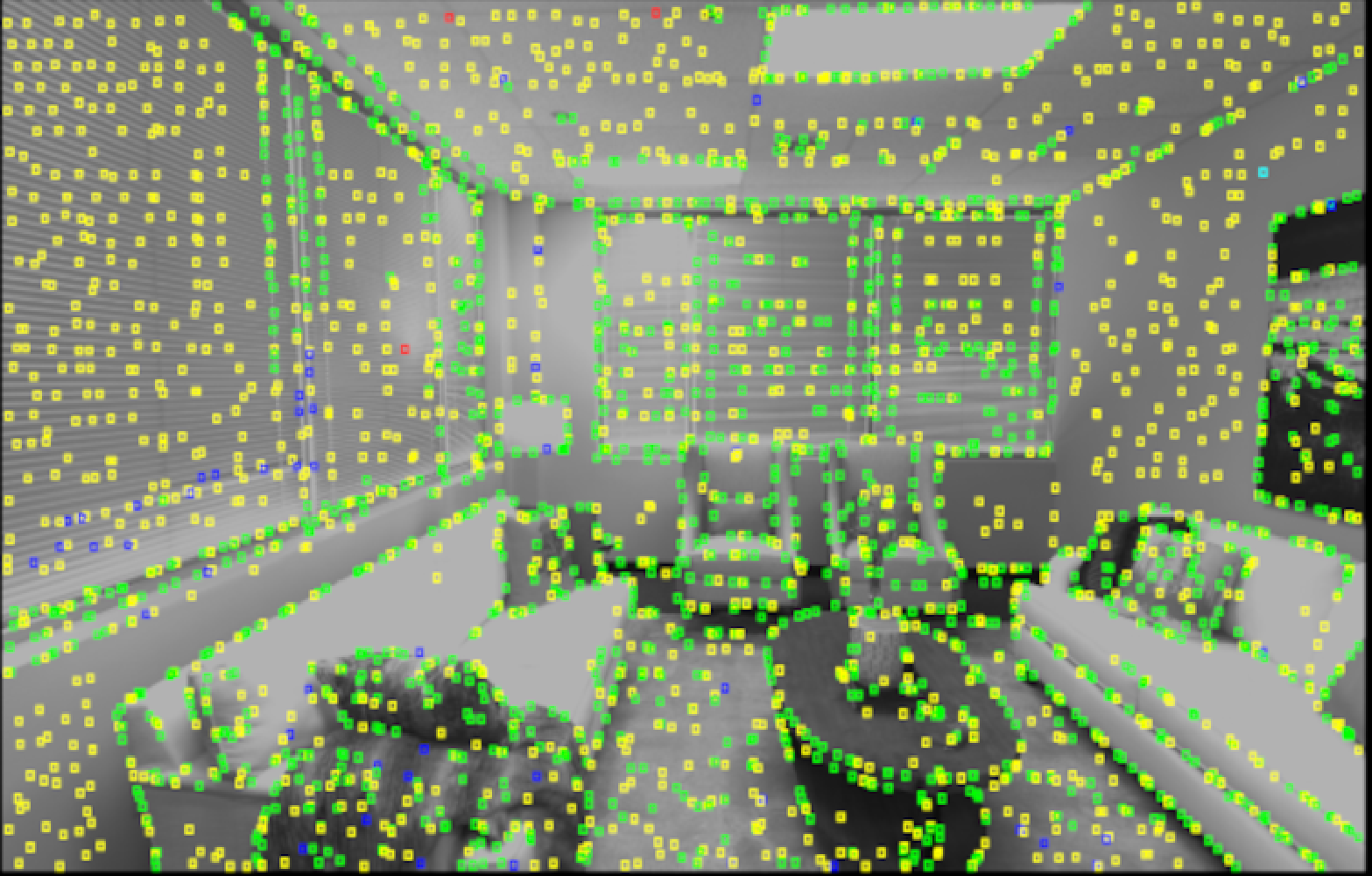}
\end{subfigure}
\caption{Tracked points from ORBSLAM3 (left) compared to our system (right). Our system tracks much more points than ORBSLAM3, which results in denser point clouds outputs.}
\label{fig:tracked}
\end{figure}

\begin{figure}[thp]
\vspace{0.2cm}
\centering
\begin{subfigure}{.25\textwidth}
  \centering
  \includegraphics[width=0.69\linewidth]{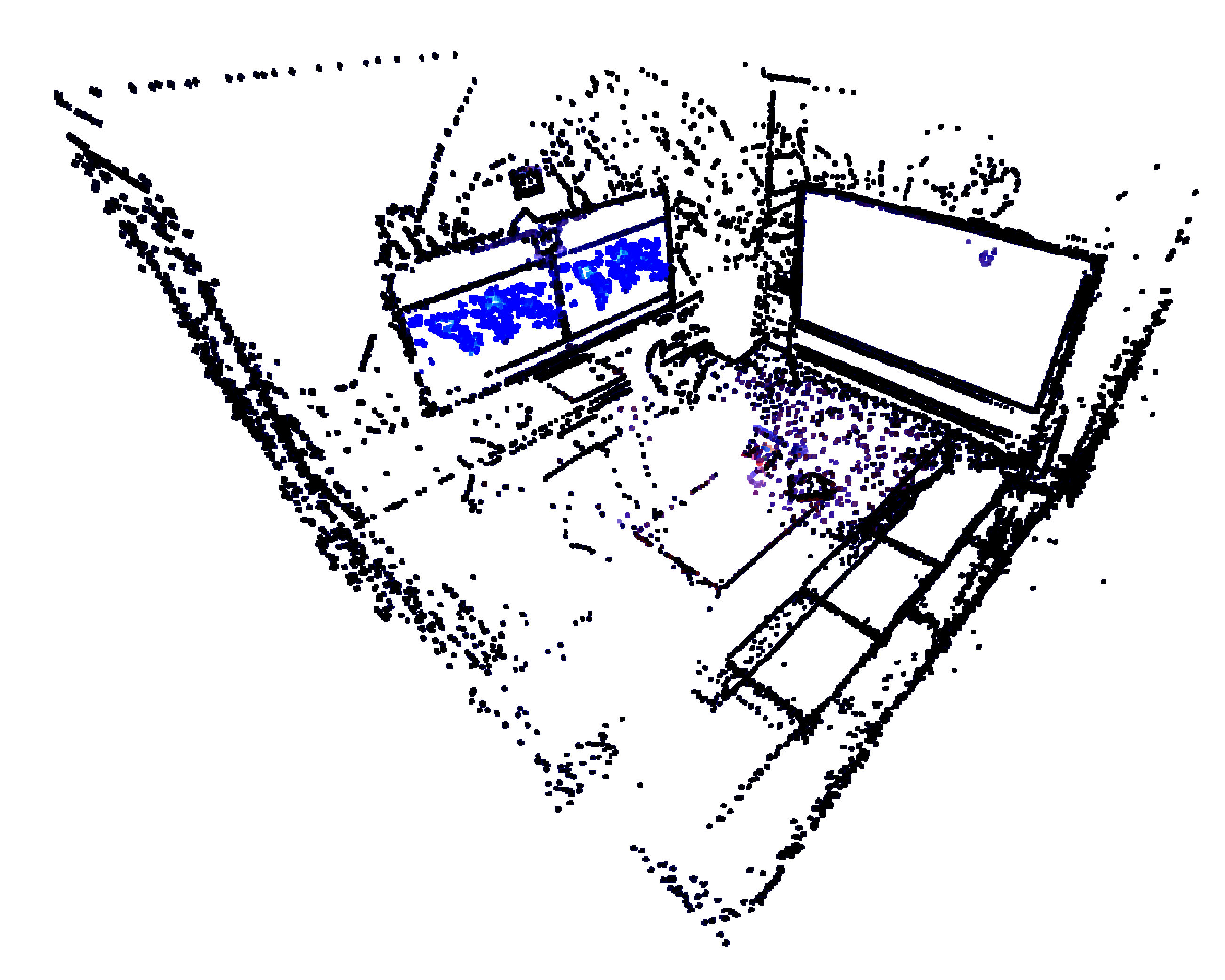}
\end{subfigure}%
\begin{subfigure}{.25\textwidth}
  \centering
  \includegraphics[width=0.69\linewidth]{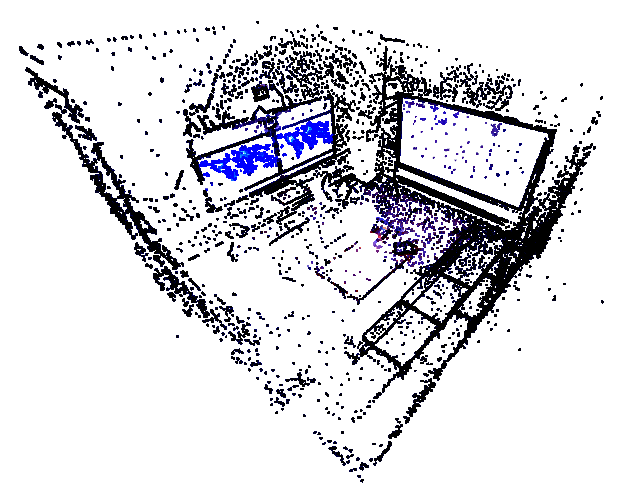}
\end{subfigure}
\caption{Original DSO Point cloud (left) compared to our system's point cloud (right). Our system has much more output points, especially in flat low-gradient regions. }
\label{fig:pointclouds}
\end{figure}

A major aspect of DSO's well-structured dense point cloud is it's pixel selection strategy. DSO does pixel selection by dividing the image into blocks and selecting the highest-gradient pixel above a gradient threshold in each block. It then repeats the process with a lower threshold and larger blocks. This approach not only tracks more pixels in complex areas but also ensures pixel selection in simpler regions. It differs from traditional methods, which typically only track easily recognizable features such as corners and edges. The differences between the two approaches can be observed in Figure \ref{fig:tracked}. This is important because we observed that while completed 3DGS maps have more Gaussians in complex areas, they still maintain some Gaussians in non-complex areas. Furthermore, DSO tracks pixels with high gradients, which are much more common than trackable feature points. Consequently, DSO's output point cloud more closely matches the density of completed 3DGS maps. 
Our experiments revealed that whileDSO's pixel selection density is optimal for tracking, increasing the pixel selection density enhances 3DGS performance, particularly in low-gradient areas that are challenging for tracking. To address this issue, we modified DSO to include additional tracked pixels not used for pose estimation to increase the output point cloud density (Figure \ref{fig:pointclouds}). This modification allows the system to have the optimal pixel density for both tracking and 3DGS. Despite these enhancements, flat regions with minimal to no gradients remain sparsely populated with tracked pixels. This is because DSO's pixel tracking system requires at least some gradient for tracking, and as a result, pixels in areas with no gradient are never tracked. 

We observed that 3DGS performs better with slightly misplaced initialized points in flat areas than with none at all. Therefore, we implemented an interpolation method that estimates point locations in low-gradient regions based on nearby tracked pixels. Our method employs the Delaunay triangulation algorithm \cite{LAWSON1972365} to divide the image into a series of triangles using tracked pixels as vertices. The depth of each interpolated point is calculated as the average depth of the triangle's vertices, which generally provides accurate results for pixels on flat surfaces. While feature-based systems like Photo-SLAM also interpolate inactive 2D feature points, our method outperforms theirs due to a higher initial point count and by focusing interpolation on flat areas where it's most accurate, which can be observed in Figure \ref{comparefigures}.

\subsection{Dense Reconstruction}
MGSO employs 3DGS as its dense reconstruction method. Following the original 3DGS, we map the scene using a set of anisotropic Gaussians \(G\). Each Gaussian \(G_{i}\) is modeled with an opacity, rotation, location, scale, and color. We follow Mono-GS' \cite{Matsuki:Murai:etal:CVPR2024} technique of representing color using RGB instead of spherical harmonics because Mono-GS showed this increased speed for minimal impact on reconstruction quality. We render RGB images of the map using the original differentiable tile-based rasterization introduced in 3DGS. 




The parameters of each Gaussian are optimized using gradient descent to minimize the photometric loss \(L\):

\begin{equation}
L = |I_r - I_{gt}| (1-\lambda) + SSIM(I_r,I_{gt}) \lambda
\end{equation}

\noindent where \(I_r\) denotes the rendered image, \(I_{gt}\) refers to the captured image, \(\lambda\) is a weighting factor and SSIM \cite{1284395} represents the structural similarity metric. 

In order to improve the speed of our system, we employ Gaussian-pyramid based learning introduced in Photo-SLAM \cite{hhuang2024photoslam} to progressively train the Gaussian map. The pyramid helps accelerate training for live video scenarios. A multi-scale Gaussian pyramid is created by repeatedly smoothing and down-sampling ground-truth image captured by the camera. The photometric loss calculation progresses from using the highest pyramid level for \(I_{gt}\) in initial iterations to lower levels as training advances. Furthermore, we use an optimized version of the 3DGS CUDA back-end \cite{Patas} that is faster than the original.

Our adaptive control strategy periodically densifies and prunes Gaussians every 1000 training iterations to improve map quality over time. We design our strategy around the point clouds returned by our SLAM module, similar to how Photo-SLAM tailored their strategy to ORBSLAM3 \cite{9440682}. The point clouds generated by our SLAM system are characterized by their high density and uniform coverage. They adapt to the scene's complexity, concentrating points in intricate areas while maintaining representation in simpler regions. When a novel keyframe is processed, we initialize new Gaussians with location and color taken from the point cloud created by the SLAM system.

However, we noticed the emergence of floaters when utilizing the 3DGS's densification and pruning strategies. To mitigate the presence of floaters, we utilized the adaptive control strategies in AbsGS \cite{ye2024absgs}. Thus, as part of our adaptive control strategy we periodically densify Gaussians with high homo-directional view-space position gradients by splitting or cloning them. Large, high-variance Gaussians are split, while small Gaussians in under-reconstructed regions are cloned.  Furthermore, we periodically prune Gaussians with low opacity to remove transparent floaters. We use the same splitting and cloning parameters as original 3DGS.

\section{Experiments and Discussions}



We evaluate MGSO against the latest state-of-the-art 3DGS dense SLAM systems: MonoGS \cite{Matsuki:Murai:etal:CVPR2024}, GlORIE-SLAM \cite{sandström2024splatslamgloballyoptimizedrgbonly}, Splat-SLAM \cite{keetha2024splatam}, IG-SLAM \cite{sarikamis2024igslam}, and Photo-SLAM \cite{hhuang2024photoslam}, to demonstrate our system's  combination of high-quality reconstruction, efficient runtime, and compact maps.

\subsection{Implementation and Setup}

\textbf{Datasets:} Evaluations are done on the sythetic Replica \cite{replica19arxiv} dataset and real-life EuRoC MAV \cite{doi:10.1177/0278364915620033} and TUM-RGBD \cite{sturm12iros} datasets. These datasets are commonly used to evaluate dense SLAM systems. 

\textbf{Hardware:} Results for Replica and re-testing Photo-SLAM were done on an Intel i9-14900K with a NVIDIA RTX 4090. The laptop runs for Replica were done on an Intel i7-12700H with a NVIDIA GeForce RTX 3080 mobile. The results for EuRoC and TUM were done on an Intel i5-12600KF with a NVIDIA RTX 3090. 

\begin{table*}[htbp]
\renewcommand{\arraystretch}{1.1}
\vspace{0.2cm}
\caption{Reconstruction Results on Replica (cm)}
\begin{center}
\begin{tabular}{l c |c c c c c c c c | c | c c c}
\hline
\textbf{Method} &  \textbf{metric} & \textbf{o0} & \textbf{o1} & \textbf{o2} & \textbf{o3} & \textbf{o4} & \textbf{r0} & \textbf{r1} & \textbf{r2} & \textbf{Avg.} & \textbf{Map Size} & \textbf{FPS} \\
\hline 
\multirow{3}{*}{Photo-SLAM} & PSNR[dB] ↑ & 35.22 & 34.35 & \textbf{29.58} & 28.55 & \textbf{32.05} & 26.75 & 27.78 & 29.43 & 30.46 & \multirow{3}{*}{22.5 MB} & \multirow{3}{*}{\textgreater 30} \\
 & SSIM ↑ & 0.94 & 0.93 & 0.91 & 0.89 & 0.92 & 0.79 & 0.84 & 0.89 & 0.89 & &  \\
 & LPIPS↓  & \textbf{0.21} & \textbf{0.23 }& \textbf{0.26} & \textbf{0.26} & \textbf{0.22} & \textbf{0.31 }& 0.28 & \textbf{0.24} & \textbf{0.25 }& & \\
 \hline
\multirow{3}{*}{MGSO} & PSNR[dB] ↑ & 35.85 & 37.15 & 29.19 & \textbf{30.44 }& 30.08 & 27.71 & 29.50 & 31.33 & 31.41 & \multirow{3}{*}{\textbf{4.6 MB}} & \multirow{3}{*}{30}  \\
 & SSIM ↑ & 0.94 & 0.94 & 0.90 & 0.90 & 0.91 & 0.79 & 0.86 & 0.91 & 0.89 & & \\
 & LPIPS↓  & 0.22 & 0.25 & 0.29 & \textbf{0.26} & 0.26 & 0.33 &\textbf{ 0.27 }&\textbf{ 0.24 }& 0.27 & & \\
 \hline
\multirow{3}{*}{MGSO (laptop)} & PSNR[dB] ↑ & \textbf{36.34} & \textbf{38.20 }& 28.90 & 30.27 & 31.41 & \textbf{28.11} & \textbf{30.04} & \textbf{31.89} & \textbf{31.90} & \multirow{3}{*}{5.2 MB} & \multirow{3}{*}{30}  \\
 & SSIM ↑ & \textbf{0.95} & \textbf{0.96 }& 0.90 & \textbf{0.91 }& \textbf{0.93 }& \textbf{0.82} & \textbf{0.87} &\textbf{ 0.92} & \textbf{0.91} & & \\
 & LPIPS↓  & 0.24 & 0.25 & 0.31 & 0.27 & 0.25 & 0.35 & 0.29 & 0.26 & 0.28 & & \\
\hline 
\end{tabular}
\label{table:replica:reconstruction}
\end{center}
\end{table*}

\textbf{Experimental Setup:} 
We utilize the default optimization configuration of 3DGS with the exception of adjusting the densification interval to 1000.
We configure the SLAM module (DSO \cite{dso:engel18}) to the default tracking settings.
The parameters for increasing SLAM output point density through the inclusion of untracked points were determined through iterative testing.

For our experiments on the EuRoC MAV dataset, we implemented a preprocessing step involving undistorting and cropping the images before inputting them into the SLAM systems. This procedure was necessary to resolve the challenge of aligning poses between the undistorted 3DGS map and the distorted ground truth images. We also re-evaluated Photo-SLAM using this modified dataset, and notably, our new tests showed significant improvements to previously reported performance (Table \ref{tableEuroc}).

\textbf{Evaluation:} To ensure a fair comparison, we evaluated our system by inputting the output 3DGS maps and pose estimation data into the original 3DGS rendering and metric scripts. We evaluate our reconstructions with the standard image quality metrics: PSNR, SSIM \cite{1284395}, and LPIPS \cite{zhang2018perceptual}.
Using an third-party evaluation system rather than built-in metrics offers a more realistic assessment, accounting for real-world factors like potential misalignment between the poses and map. Consistent with other dense SLAM systems, we evaluate on every fifth frame. Photo-SLAM's evaluations are updated using this methodology to ensure consistency. We did ten runs for the Replica and EuRoC dataset and five runs for the TUM-RGBD dataset. Results for other systems were obtained from their respective publications, with the exception of Mono-GS, whose results were sourced from Splat-SLAM \cite{sandström2024splatslamgloballyoptimizedrgbonly}. 

However, the evaluation process tends to favor slower systems, as 3DGS performs better with extended training times, which may create bias against faster systems. Therefore, readers should consider the differences in speed when interpreting results. Because our system inherits real-time constraint handling from DSO, we decided to constrain our speed to the real-time speeds of videos to enhance the realism of the results. Replica and TUM-RGBD were run at 30 fps while EuRoC was run at 20 fps.

\subsection{Discussion}

\subsubsection{Localization}

\begin{table}[htbp]
 \renewcommand{\arraystretch}{1.1}
 \setlength{\tabcolsep}{3.6pt}
 \caption{Absolute Trajectory Error of Tracking on Replica (RMSE in cm)}
 \begin{center}
 \begin{tabular}{l|c c c c c c c c | c}
 \hline
 \textbf{Method} & \textbf{r0} & \textbf{r1} & \textbf{r2} & \textbf{o0} & \textbf{o1} & \textbf{o2} & \textbf{o3} & \textbf{o4} & \textbf{Avg.} \\
 \hline 
 Photo-SLAM & 0.58 & \textbf{0.32} & \textbf{5.03} & 0.47 & 0.58 & 0.35 & 1.18 & 0.23 & \textbf{1.09}\\
 MGSO & \textbf{0.35} & 1.02 & 5.93 & \textbf{0.22} &\textbf{ 0.54} & \textbf{0.28} & \textbf{0.34} & \textbf{0.2} & 1.11\\
 \hline 
 \end{tabular}
 \label{tableTrack}
 \end{center}
 \end{table}

While we include tracking results in Table \ref{tableTrack}, tracking performance is not our system's focus. We did not modify the localization aspect of DSO and should inherit its performance. Our system's comparable tracking to Photo-SLAM suggests any rendering differences are not due to localization.

\subsubsection{Reconstruction Quality}
\begin{figure*}%
\vspace{0.2cm}
\begin{subfigure}{.195\textwidth}
  \centering
  \includegraphics[width=1.0\linewidth]{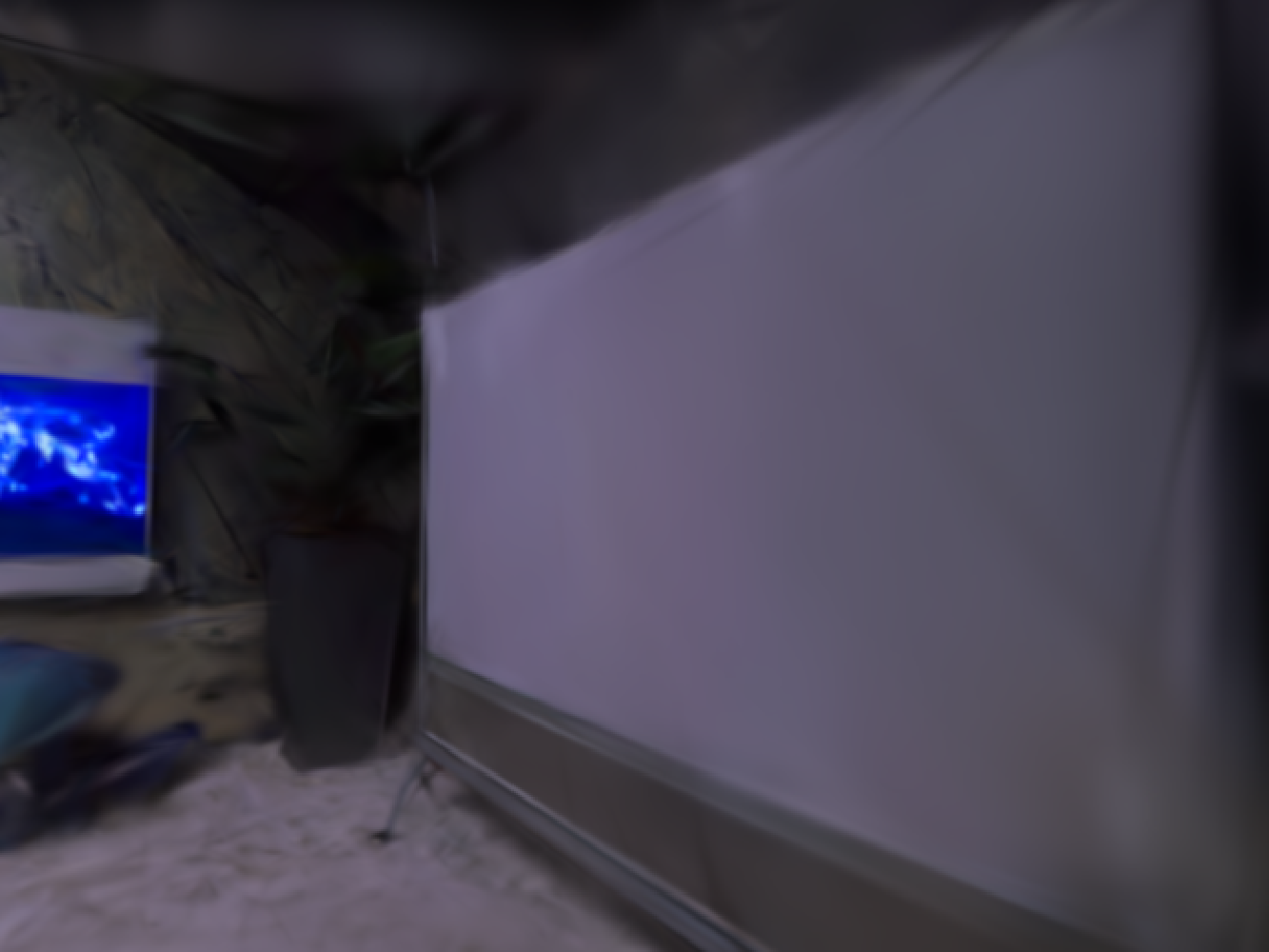}
\end{subfigure}%
\hspace*{\fill}
\begin{subfigure}{.195\textwidth}
  \centering
  \includegraphics[width=1.0\linewidth]{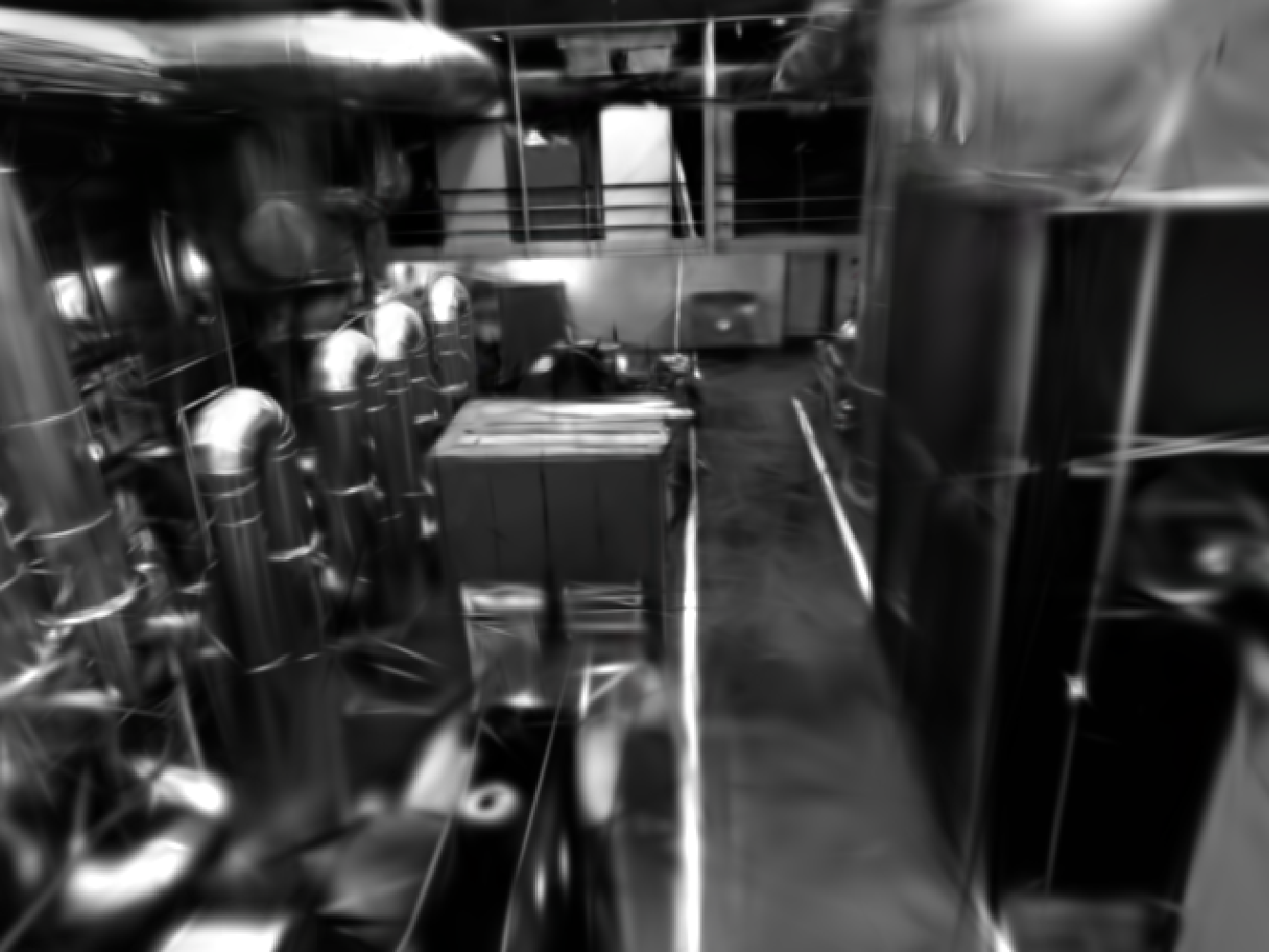}
\end{subfigure}%
\hspace*{\fill}
\begin{subfigure}{.195\textwidth}
  \centering
  \includegraphics[width=1.0\linewidth]{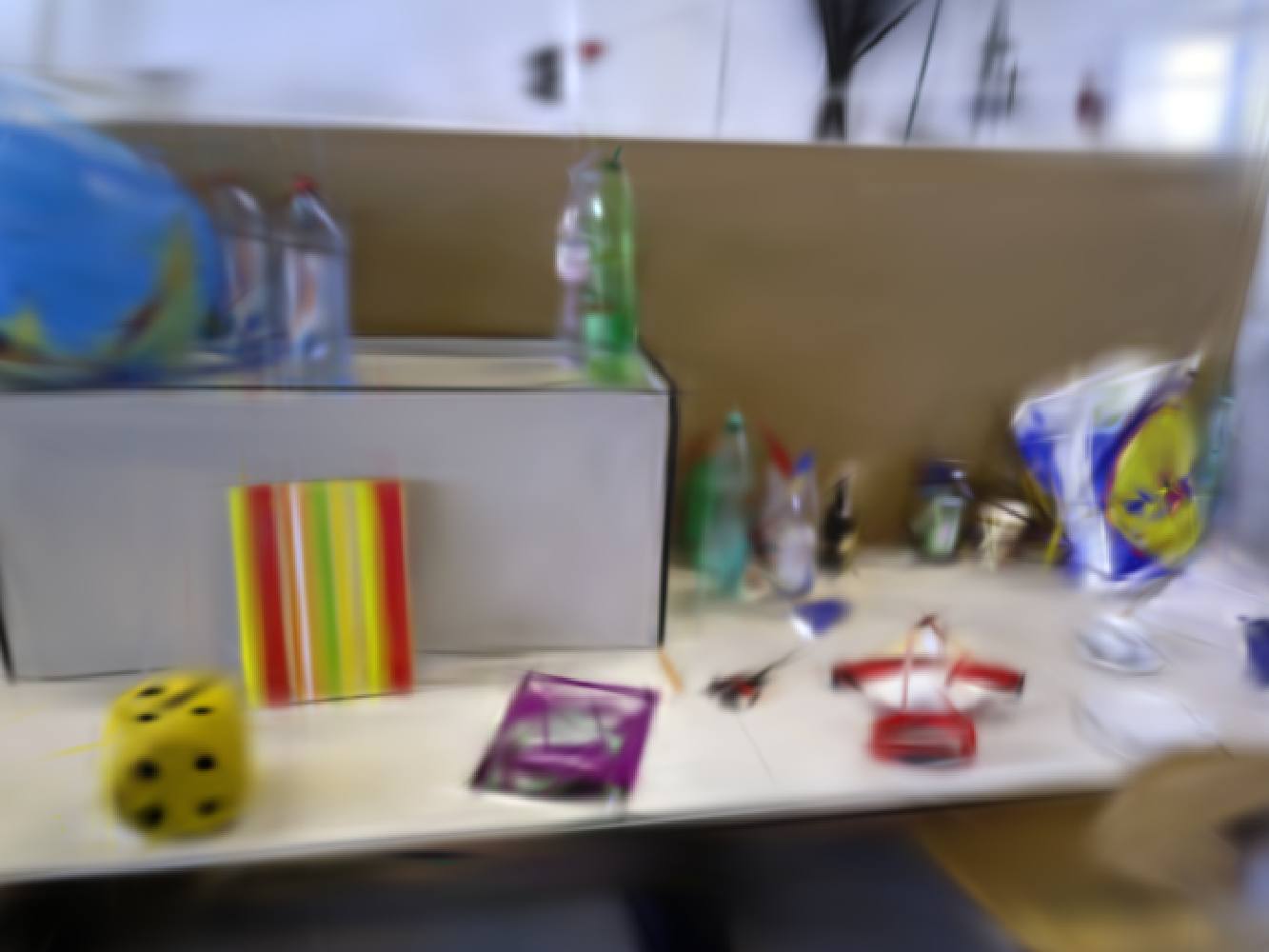}
\end{subfigure}%
\hspace*{\fill}
\begin{subfigure}{.195\textwidth}
  \centering
  \includegraphics[width=1.0\linewidth]{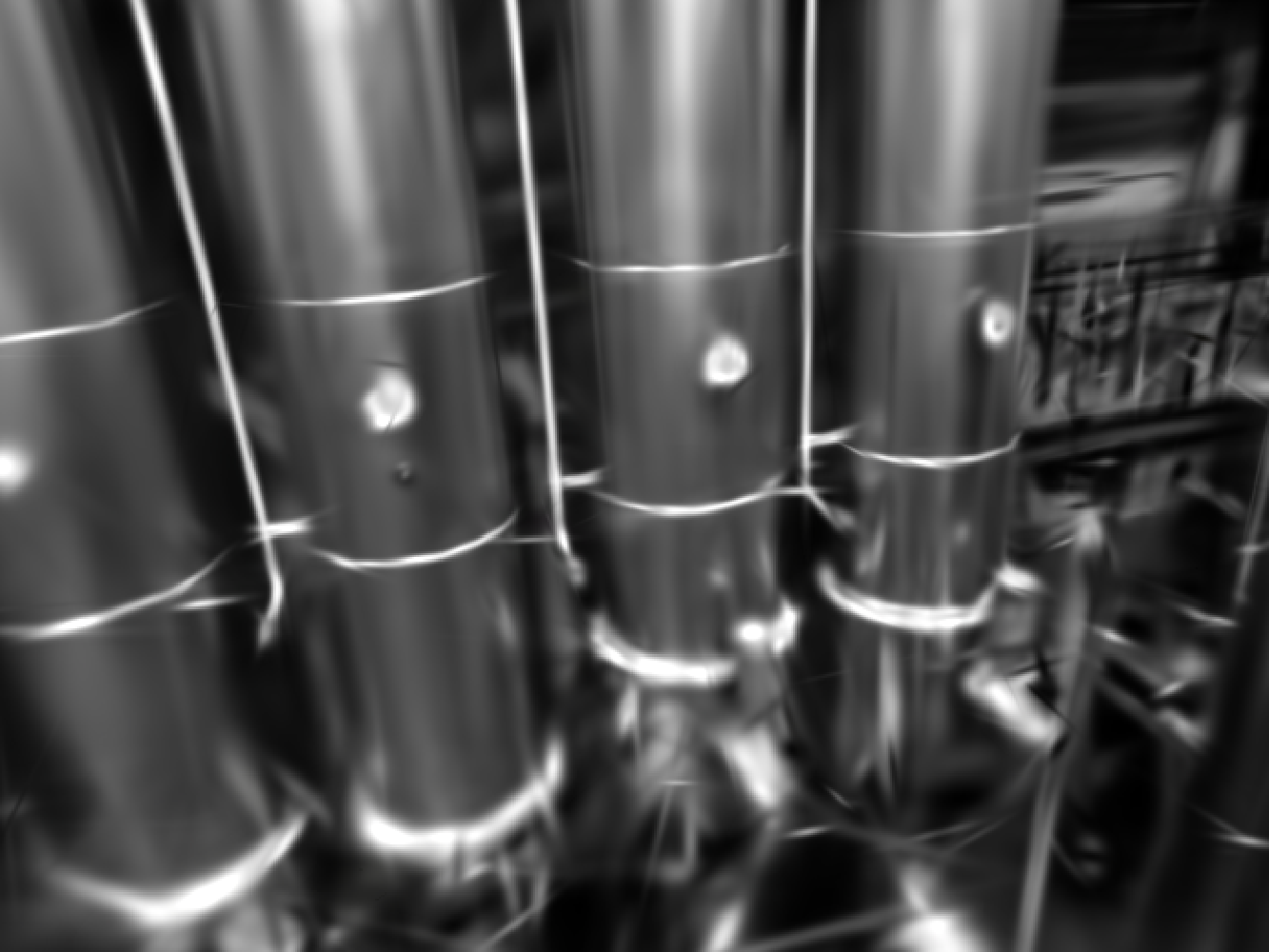}
\end{subfigure}%
\hspace*{\fill}
\begin{subfigure}{.195\textwidth}
  \centering
  \includegraphics[width=1.0\linewidth]{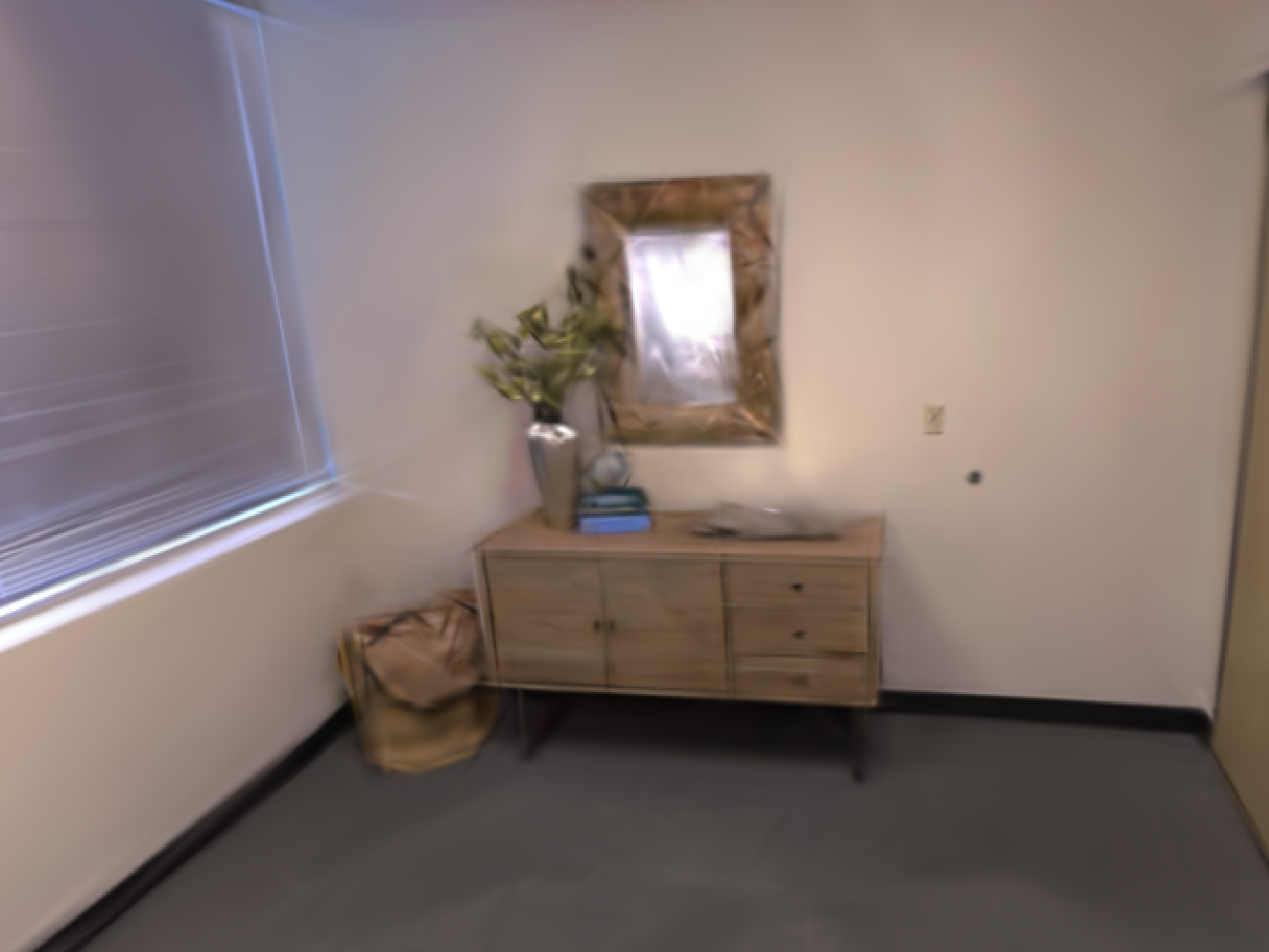}
\end{subfigure}%

\begin{subfigure}{.195\textwidth}
  \centering
  \includegraphics[width=1.0\linewidth]{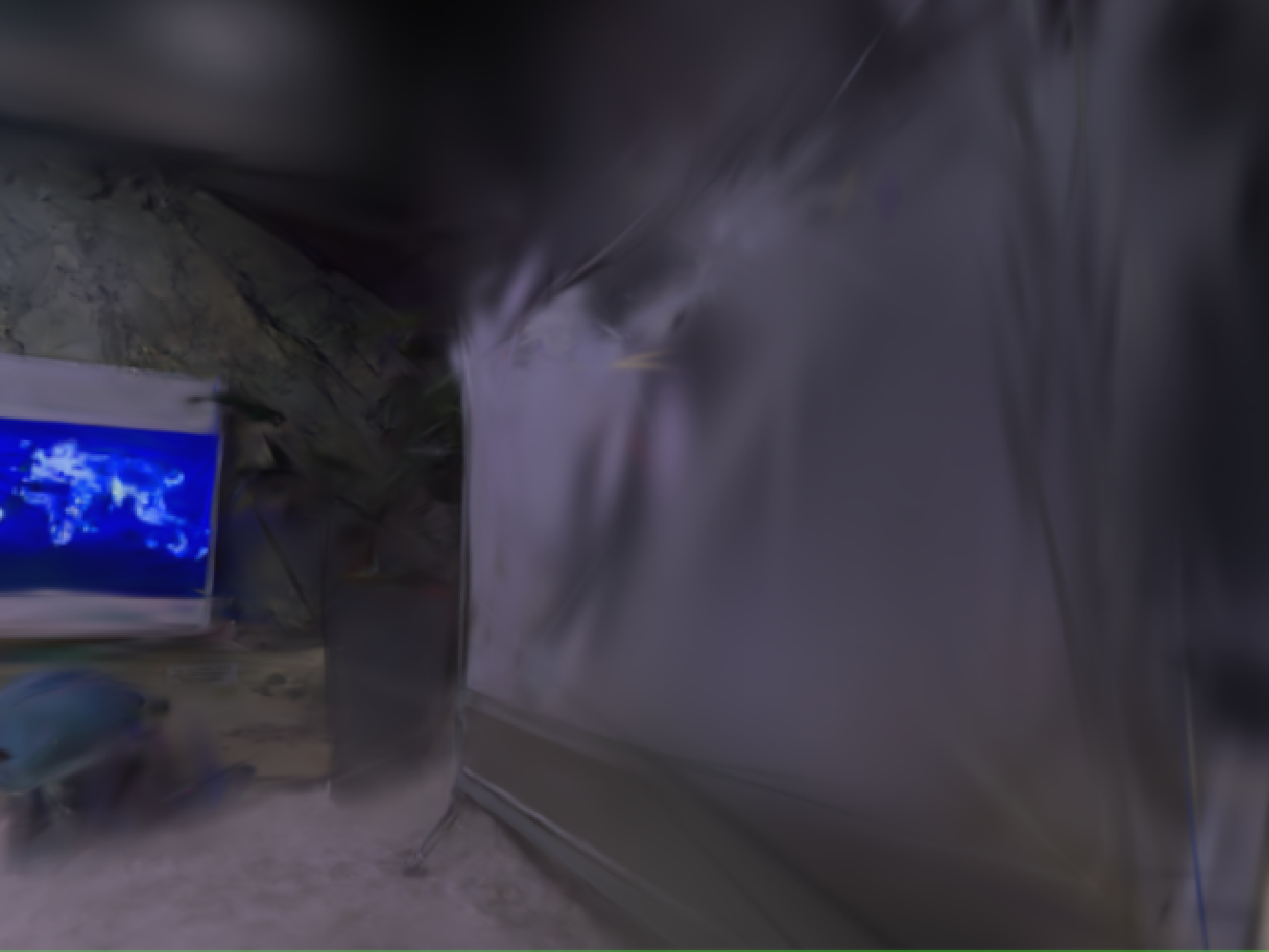}
  \caption{MGSO renders flat \\surfaces well}
\end{subfigure}%
\hspace*{\fill}
\begin{subfigure}{.195\textwidth}
  \centering
  \includegraphics[width=1.0\linewidth]{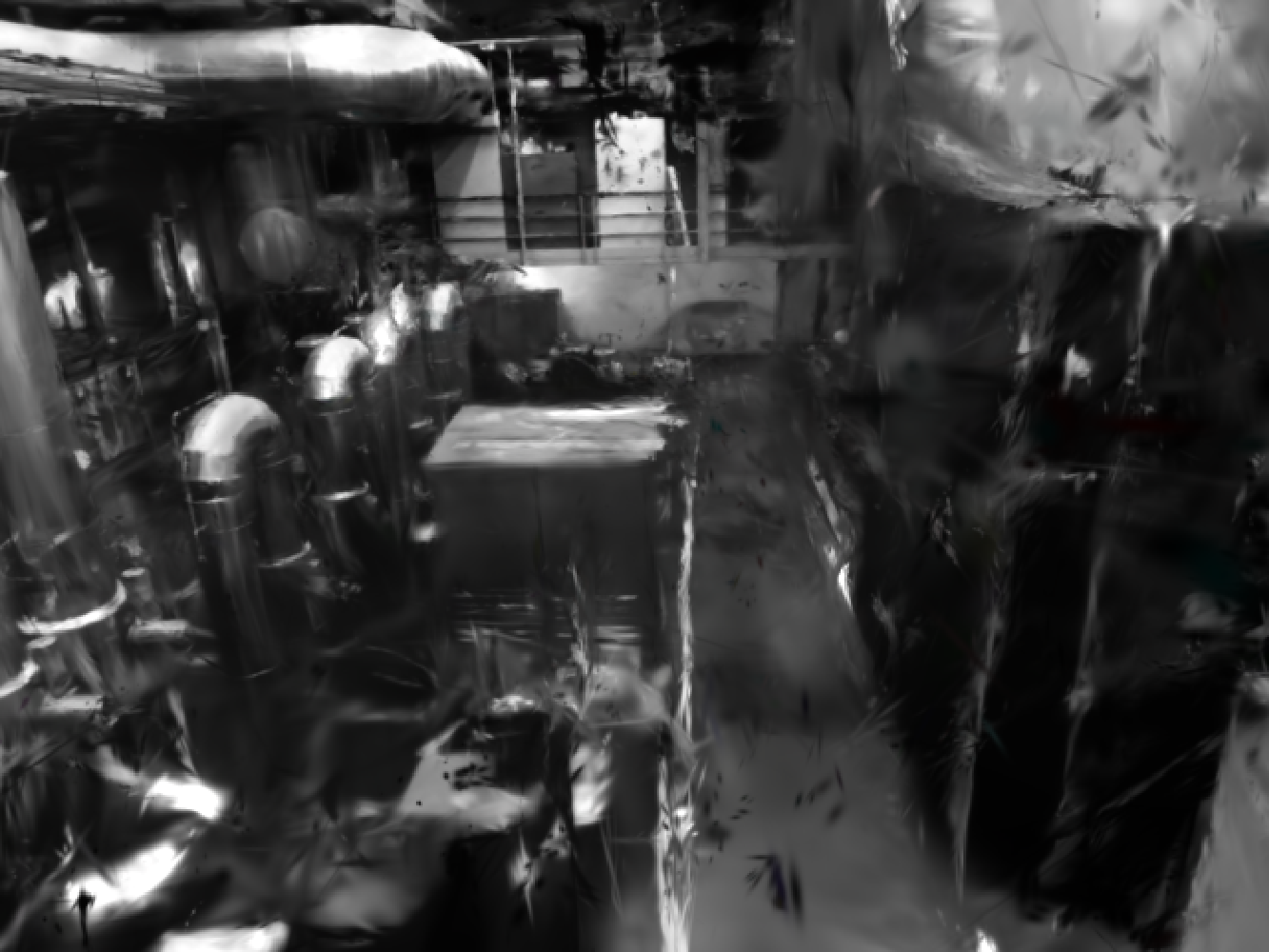}
\caption{MGSO has less floaters and artifacts}
\end{subfigure}%
\hspace*{\fill}
\begin{subfigure}{.195\textwidth}
  \centering
  \includegraphics[width=1.0\linewidth]{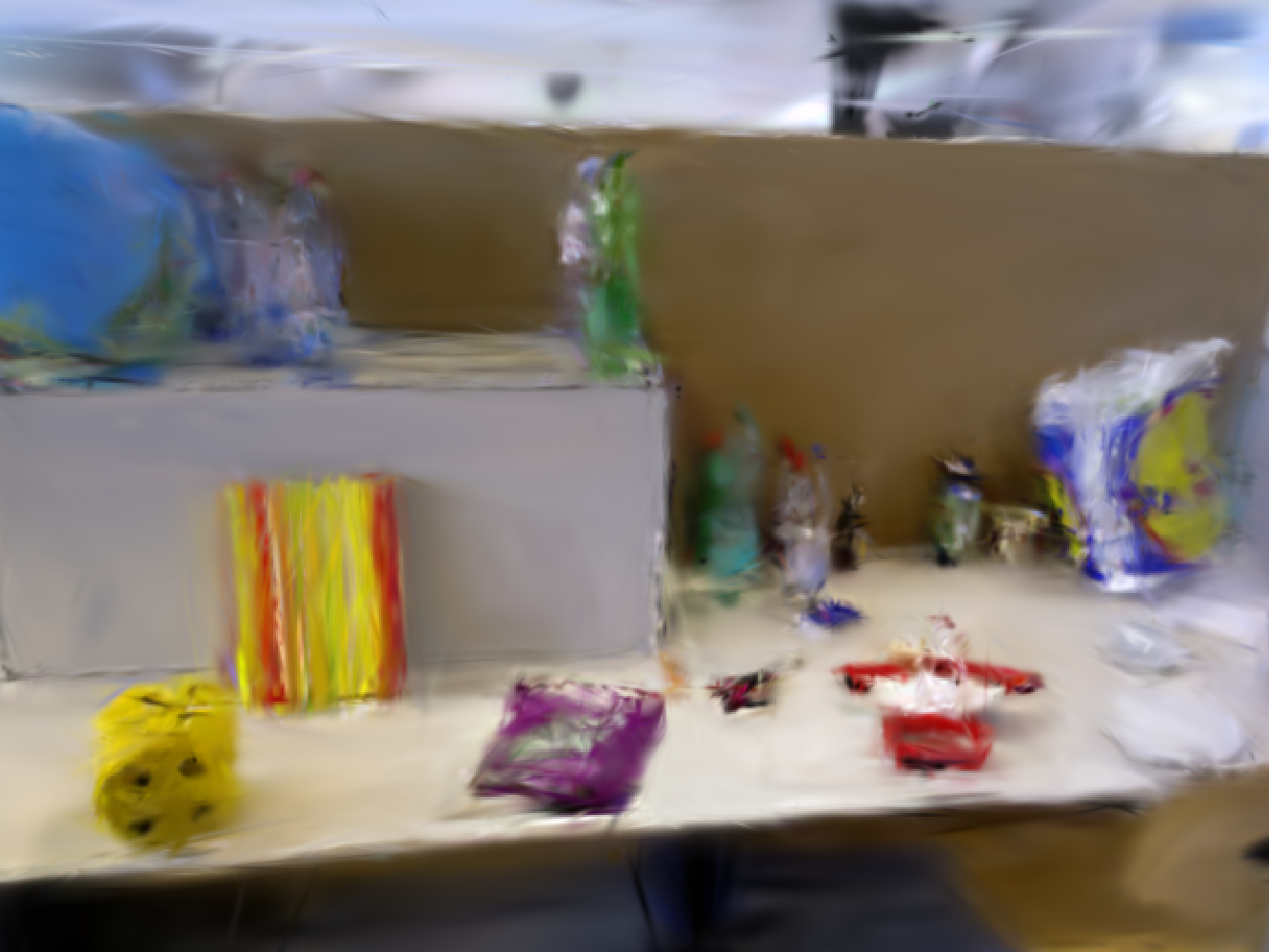}
\caption{MGSO has better edges on difficult scene}
\end{subfigure}%
\hspace*{\fill}
\begin{subfigure}{.195\textwidth}
  \centering
  \includegraphics[width=1.0\linewidth]{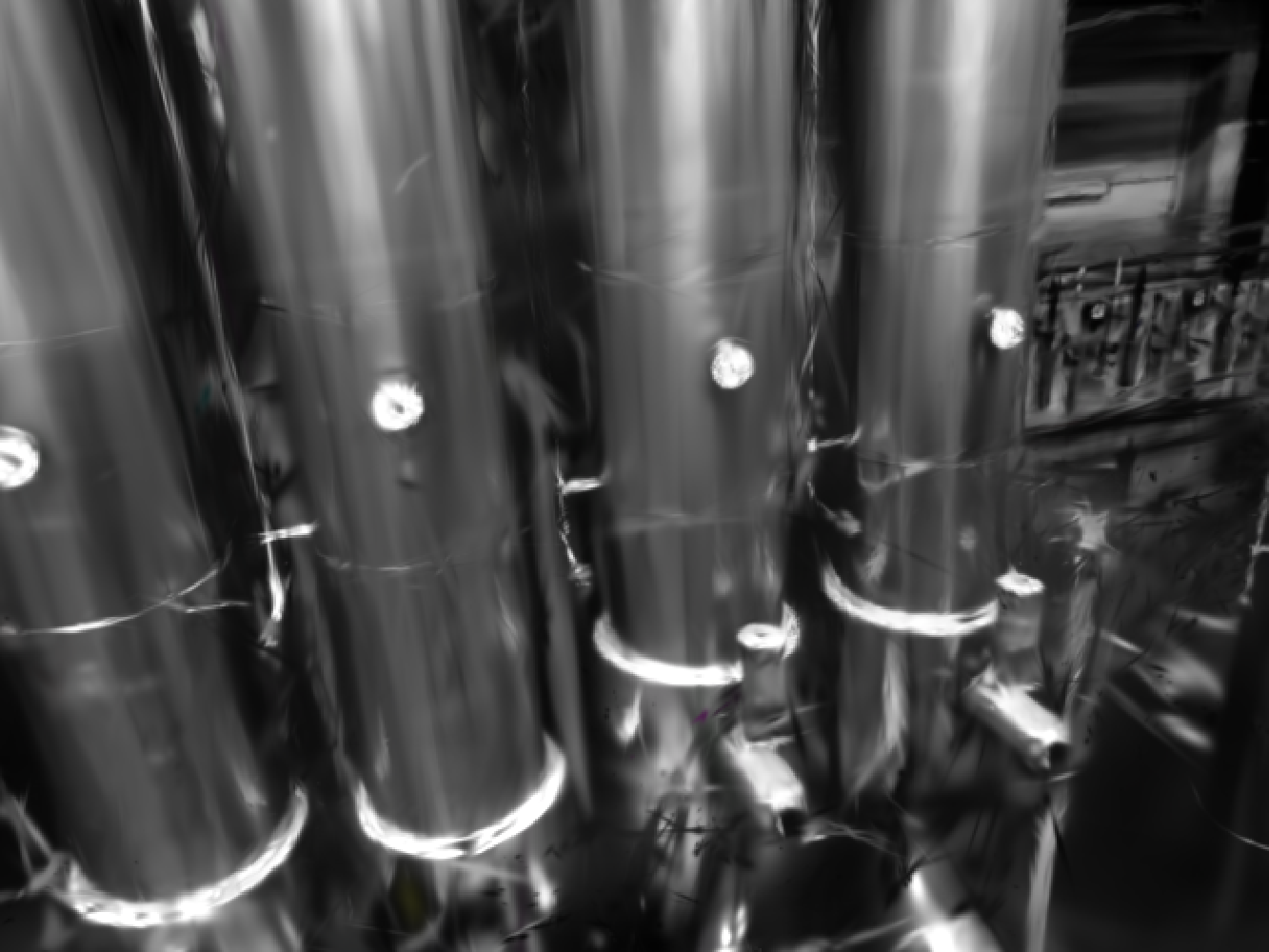}
\caption{MGSO renders thin features better}
\end{subfigure}%
\hspace*{\fill}
\begin{subfigure}{.195\textwidth}
  \centering
  \includegraphics[width=1.0\linewidth]{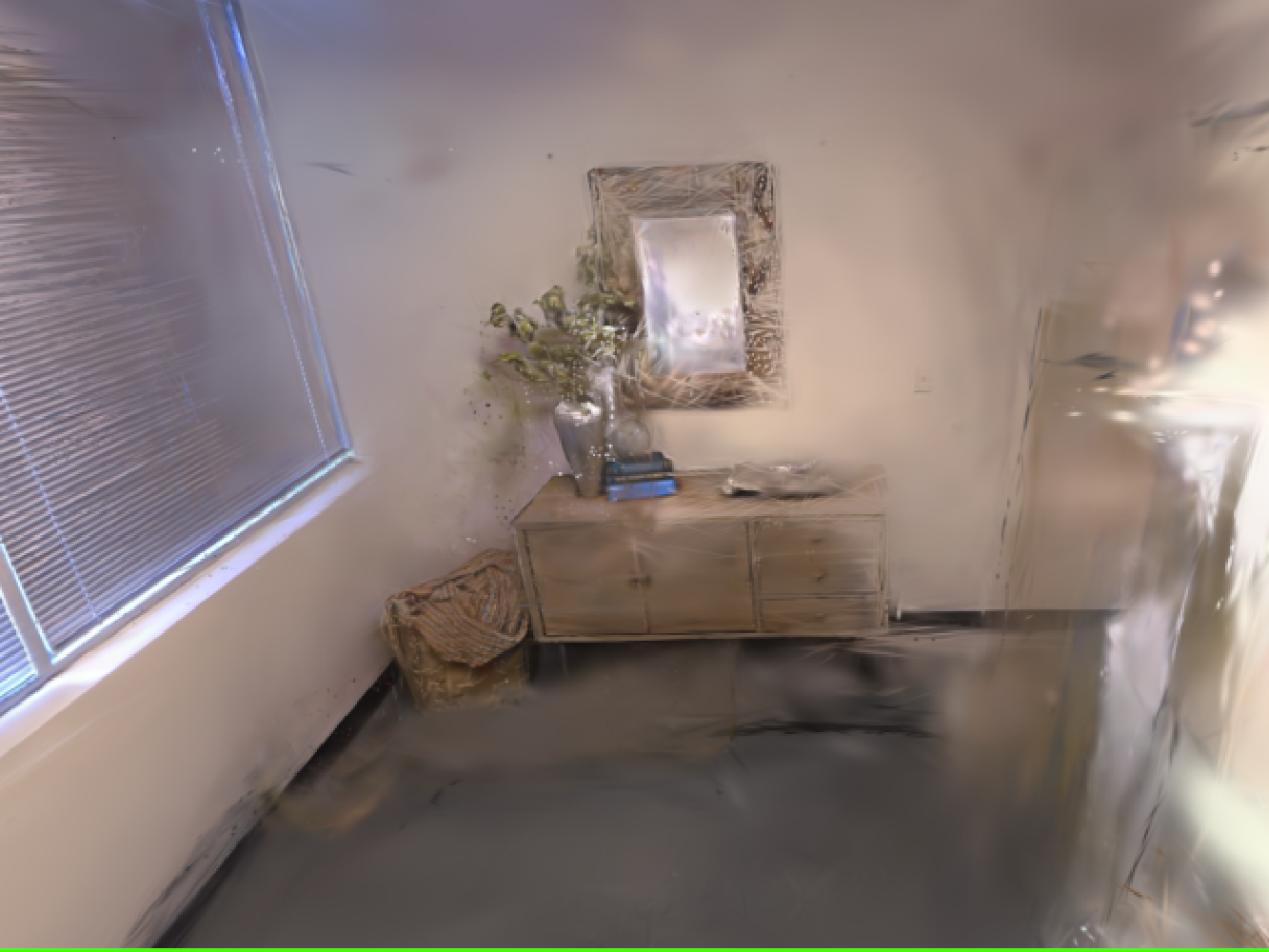}
  \caption{MGSO has less floaters and artifacts}
\end{subfigure}%
\caption{Comparison of difficult novel view renders between MGSO (top) and Photo-SLAM (bottom). Captions describe how MGSO performs better.}\label{fig:photocompare}
\end{figure*}

MGSO consistently achieves high PSNR and SSIM across all datasets. In the Replica dataset (Table \ref{table:replica:reconstruction}), MGSO outperforms Photo-SLAM with a PSNR of 31.406 dB and a much smaller map size of 4.618 Mb, with its mobile version showing even better results (31.896 dB PSNR, 0.906 SSIM). On the EuRoC dataset (Table \ref{tableEuroc}), MGSO further demonstrates superior performance with 22.10 dB PSNR and 0.80 SSIM, compared to Photo-SLAM's 19.68 dB and 0.75 SSIM. Similar trends are observed on the TUM dataset (Table \ref{tableTUM}), where MGSO achieves a higher PSNR and SSIM than Photo-SLAM.
MGSO’s key advantage is its ability to generate dense, well-structured point clouds, requiring less refinement and resulting in more compact maps—half the size of Photo-SLAM’s. This efficient initialization reduces the need for extensive operations like cloning and pruning, leading to faster convergence and fewer reconstruction artifacts. Figure \ref{fig:photocompare} further highlights MGSO’s improved rendering of flat surfaces, fewer floating artifacts, and better preservation of edges and thin features, showcasing its capacity to handle complex scenes with more accurate and detailed reconstructions.
\begin{table}[htbp]
\renewcommand{\arraystretch}{1.1}
\caption{Reconstruction Results on EuRoC}
\begin{center}
\begin{tabular}{l c |c c c | c | c }
\hline
\textbf{Method} &  \textbf{metric} & \textbf{MH} & \textbf{V1} & \textbf{V2} & \textbf{Avg.} & \textbf{Mem.} \\
\hline 
 \multirow{3}{*}{\parbox{1cm}{Photo-SLAM}} & PSNR[dB] ↑ & 18.60 & 18.30 & 17.94 & 18.28 & \multirow{3}{*}{111.8} \\
 & SSIM ↑ & 0.65 & 0.73 & 0.65 & 0.68 &  \\
 & LPIPS↓  & 0.39 & 0.44 & 0.53 & 0.46  & \\
 \hline
 \multirow{3}{*}{MGSO} & PSNR[dB] ↑ & \textbf{20.75} & \textbf{20.26} & \textbf{20.31}& \textbf{20.44} & \multirow{3}{*}{\textbf{8.3}} \\
 & SSIM ↑ & \textbf{0.72} & \textbf{0.79} & \textbf{0.75}  & \textbf{0.76 }&  \\
 & LPIPS↓  & \textbf{0.36} & \textbf{0.39 }& \textbf{0.39} & \textbf{0.38 }& \\
 \hline
 \multicolumn{7}{l}{Mem. is average map size in Mb}
\end{tabular}
\label{tableEuroc}
\end{center}
\end{table}

\begin{table}[htbp]
\renewcommand{\arraystretch}{1.1}
\caption{Reconstruction Results on TUM's}
\begin{center}
\begin{tabular}{l c |c c c | c }
\hline
\textbf{Method} &  \textbf{metric} & \textbf{fr1} & \textbf{fr2} & \textbf{fr3} & \textbf{Avg.} \\
\hline 
\multirow{3}{*}{Photo-SLAM} & PSNR[dB] ↑ & 18.01 & 16.93 & 17.11& 17.35  \\
 & SSIM ↑ & 0.65 & 0.60 & 0.62  & 0.63 \\
 & LPIPS↓  & \textbf{0.41} & 0.41 & 0.42 & 0.41 \\
 \hline
 \multirow{3}{*}{MGSO} & PSNR[dB] ↑ & \textbf{18.07} & \textbf{24.10} & \textbf{21.61} & \textbf{21.26} \\
 & SSIM ↑ & \textbf{0.66} & \textbf{0.80} & \textbf{0.75} &\textbf{ 0.74}  \\
 & LPIPS↓  & 0.45 & \textbf{0.33 }& \textbf{0.38} &\textbf{ 0.39} \\
 \hline
\end{tabular}
\label{tableTUM}
\end{center}
\end{table}
\subsubsection{Resource Efficiency and Real-Time Performance}
\begin{figure}[htbp]
\vspace{0.2cm}
\begin{tikzpicture}

\pgfplotsset{
  set layers,
  mark layer=axis tick labels
}

\begin{axis}[
height=110, width = 245,
xmin=1, xmax=150,xmode=log,
ymin=25, ymax=40, 
xlabel={Map Size (Mb)}, 
ylabel={PSNR (dB)},
    label style={font=\bfseries\boldmath},
    tick label style={font=\bfseries\boldmath},
    scatter/classes={
        a={mark=square*, red}, 
        b={mark=diamond*, orange}, 
        c={mark=triangle*, green}
    },
legend style={at={(0.0,1.0)}, anchor=north west}
] 
\small
\addplot[scatter, only marks,
        scatter src=explicit symbolic,
        nodes near coords*={\annotvalue},
        node near coord style={rotate=90, anchor=east, font=\scriptsize},
        visualization depends on={value \thisrow{annotation} \as \annotvalue},]
table[meta=label] {
y x label annotation
6.8 36.45 a SpS
14.8 36.21 b IGS
114 31.04 a GlS
4.3 31.406 c MGSO
22.5 30.464 c PhS
6.8 31.22 a MGS
};
\legend{x\textless5,5\(\leq\)x\textless24,x\(\geq\)24}
\end{axis}
\end{tikzpicture}
\caption{Plot of Table \ref{table:replica:aggregat}. The 'x' in the legend represents the frames per second. We consider fps\textgreater24 (cinema fps standard) as real-time.}
\label{plotPSNRMap}
\end{figure}
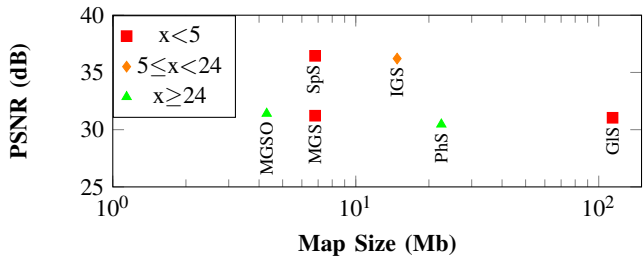

 MGSO excels in low memory usage and real-time FPS. On the EuRoC dataset (Table \ref{tableEuroc}), MGSO requires only 8.32 MB, significantly less than Photo-SLAM’s 109.73 MB, and just 2.85 MB on the TUM dataset (Table \ref{tableTUM}), compared to Photo-SLAM’s 17 MB. All the while, MGSO maintains real-time performance (Tables \ref{table:replica:reconstruction},\ref{table:replica:aggregat}) MGSO’s structured point clouds allow it to create compact maps with minimal redundant elements, resulting in lower memory consumption. This contrasts with Photo-SLAM’s larger map sizes, which require more refinement. Figure \ref{plotPSNRMap}  underscores MGSO’s balance of high FPS with low map size, we are the only system capable of real-time performance with compact maps.
 
\begin{table}[htbp]
\renewcommand{\arraystretch}{1.1}
\caption{Replica Aggregated Results}
\setlength{\tabcolsep}{3.75pt}
\begin{center}
\begin{tabular}{l | c c c c c}
\hline
\textbf{Method} & \textbf{PNSR[dB]} & \textbf{Map Size} & \textbf{FPS} & \textbf{GPU Usage} \\
\hline 
GlORIE-SLAM (GlS) & 31.04 & 114 Mb & 0.23 & 15.22 \\
Mono-GS (MGS) & 31.22 & 6.8 Mb & 0.32 & 14.62 \\
Splat-SLAM (SpS) & \textbf{36.45} & 6.8 Mb & 1.24 & 17.57 \\
IG-SLAM (IGS) & 36.21 & 14.8 Mb & 9.94 & 16.20 \\
Photo-SLAM (PhS) & 30.46 & 22.5 Mb & \textbf{\textgreater30*} & \textbf{3.62} \\
MGSO (MGSO) & 31.41 & \textbf{4.3 Mb}& \textbf{30*} & 7.98 \\
\hline
 \multicolumn{5}{l}{*System processed data as fast as inputted video stream}
\end{tabular}
\label{table:replica:aggregat}
\end{center}
\end{table}

%

\subsection{Ablations}
ted experiments to evaluate the robustness of our system to the frequency of densification and pruning. As shown in table \ref{tableDenseAblations}, increasing the rate of densification does not improve reconstruction results and instead reduces the compactness of the final 3DGS map. In fact, at high densification rates, the reconstruction quality diminishes. The observed robustness suggests that our system generates spatially accurate point clouds that effectively capture both complex and simple areas of the scene, without requiring significant adjustments from densification or pruning. 

Table \ref{table:tableInit} demonstrates the importance of dense, well-structured inputs to 3D Gaussians. We deliberately reduced the density of our point clouds by utilizing only half of the points from our tracking system and excluding additional untracked points. This intentional sparsification resulted in a marked decline in reconstruction quality.

\begin{table}[!htbp]
\setlength{\tabcolsep}{4.20pt}
\renewcommand{\arraystretch}{1.1}
\caption{Densify Iteration Ablation}
\begin{center}
\begin{tabular}{l | c c c c c c c}
\hline
\textbf{Scene} &  \textbf{metric} & \textbf{1024} & \textbf{512} & \textbf{256} & \textbf{128} & \textbf{64} & \textbf{32} \\
\hline 
\multirow{2}{*}{o0} & PSNR[dB]↑ & 37.06 & 37.10 & 37.02 & 37.15 & 37.40 & 36.11\\
 & Memory(Mb) & 6.4 & 7.2 & 9.0 & 13.0 & 21.4 & 38.1\\
\multirow{2}{*}{r0} & PSNR[dB]↑ & 28.84 & 28.73 & 28.86 & 28.76 & 28.49 & 27.73\\
 & Memory(Mb) & 4.5 & 5.3 & 7.7 & 12.4 & 22.2 & 38.5\\
\hline 
  \multicolumn{8}{l}{Ablation Evaluations done on training images instead of test images}
\end{tabular}
\label{tableDenseAblations}
\end{center}
\end{table}
\begin{table}[!htbp]
\renewcommand{\arraystretch}{1.1}
\caption{Additional Dense Points Ablation}
\begin{center}
\begin{tabular}{l | c c c c c c }
\hline
\textbf{Dataset} & \textbf{metric} & \textbf{o0} & \textbf{o1} & \textbf{o2} & \textbf{r0} & \textbf{r1} \\
\hline 
\multirow{2}{*}{Base} & PSNR[dB]↑ & 37.06 & 38.37 & 29.81 & 28.84 & 30.68 \\
 & Memory(Mb) & 6.4 & 4.2 & 6.0 & 4.5 & 5.3  \\
 \hline 
 \multirow{2}{*}{Halved} & PSNR[dB]↑ & 33.48 & 33.93 & 27.87 & 27.30 & 29.13 \\
 & Memory(Mb)& 2.0 & 1.7 & 2.2 & 2.1 & 2.0 \\
\hline 
  \multicolumn{7}{l}{Ablation evaluations done on training images instead of test images}
\end{tabular}
\label{table:tableInit}
\end{center}
\end{table}

\section{Conclusions}
MGSO integrates real-time photometric SLAM with 3D Gaussian Splatting (3DGS) to achieve dense, high-quality, and memory efficient 3D reconstruction using only a monocular camera. Our approach addressed several challenges in order to harness the natural compatibility of these two techniques. Its proven versatility across various environments without the use of depth sensors makes it optimal for robotics, AR/VR, and digital twin applications. Future research could explore implementing loop closure for global consistency and real-time re-rendering for adaptive scene reconstruction, enhancing MGSO's precision and efficiency in complex, large-scale environments.

\addtolength{\textheight}{-2cm}   

\section*{ACKNOWLEDGMENT}
This work was supported by the Natural Sciences and Engineering Research Council (NSERC) (grant-number, www.nserc-crsng.gc.ca) and the DIDYMOS-XR Horizon Europe project (grant number 101092875–DIDYMOS-XR,www.didymos-xr.eu).


\pagebreak

\bibliographystyle{IEEEtran}
\bibliography{bib/IEEEbib}
\end{document}